\documentclass[letterpaper]{article} 
\usepackage[draft]{aaai25}  
\usepackage{times}  
\usepackage{helvet}  
\usepackage{courier}  
\usepackage[hyphens]{url}  
\usepackage{graphicx} 
\usepackage{natbib}  
\usepackage{caption} 
\usepackage{algorithm}

\usepackage{newfloat}
\usepackage{listings}
\DeclareCaptionStyle{ruled}{labelfont=normalfont,labelsep=colon,strut=off} 
\floatstyle{ruled}
\newfloat{listing}{tb}{lst}{}
\floatname{listing}{Listing}
\usepackage{algpseudocode}

\usepackage{amsmath}

\usepackage{caption}
\usepackage{subcaption}
\usepackage{multirow}
\usepackage{graphicx}
\usepackage{svg}

\usepackage{array}
\usepackage{makecell}
\usepackage{caption}
\usepackage{subcaption}
\usepackage[inline]{enumitem}

\newcommand*{\ourmodel}{nach0-pc}

\title{\ourmodel{}: Multi-task Language Model with Molecular Point Cloud Encoder}

\author{
Maksim Kuznetsov \textsuperscript{\rm 1*}, Airat Valiev \textsuperscript{\rm 2}, Alex Aliper \textsuperscript{\rm 2}, Daniil Polykovskiy \textsuperscript{\rm 1}, \\
Elena Tutubalina \textsuperscript{\rm 2}, Rim Shayakhmetov \textsuperscript{\rm 1}, Zulfat Miftahutdinov \textsuperscript{\rm 1}
}

\affiliations{
    \textsuperscript{\rm 1}Insilico Medicine Canada Inc., \textsuperscript{\rm 2}Insilico Medicine AI Ltd.\\
    \textsuperscript{*} Corresponding author: kuznetsov@insilicomedicine.com
}

\begin{document}
\maketitle
\begin{abstract}
Recent advancements have integrated Language Models (LMs) into a drug discovery pipeline. However, existing models mostly work with SMILES and SELFIES chemical string representations, which lack spatial features vital for drug discovery. Additionally, attempts to translate chemical 3D structures into text format encounter issues such as excessive length and insufficient atom connectivity information.
To address these issues, we introduce \ourmodel{}, a model combining domain-specific encoder and textual representation to handle spatial arrangement of atoms effectively. Our approach utilizes a molecular point cloud encoder for concise and order-invariant structure representation. We introduce a novel pre-training scheme for molecular point clouds to distillate the knowledge from spatial molecular structures datasets.
After fine-tuning within both single-task and multi-task frameworks, \ourmodel{} demonstrates performance comparable with other diffusion models in terms of generated samples quality across several established spatial molecular generation tasks. Notably, our model is a multi-task approach, in contrast to diffusion models being limited to single tasks. Additionally, it is capable of processing point cloud-related data, which language models are not capable of handling due to memory limitations. These lead to our model having reduced training and inference time while maintaining on par performance.
\end{abstract}

\section{Introduction}

Language Models (LMs) have shown exceptional natural language understanding \cite{devlin2019bert}, and performance across diverse natural language tasks, including language translation \cite{raffel2020t5}, question answering \cite{lewis2020bart}, code generation \cite{feng2020codebert}. LMs also show efficiency as conversational agents, engaging in meaningful dialogues \cite{brown2020gpt}. Moreover, recent research has demonstrated the capacity of LMs to integrate various data modalities, including images \cite{alayrac2022flamingo, koh2023fromage}, video \cite{lu2019vilbert, sun2019videobert}, audio \cite{ren2019fastspeech, radford2023whisper} and point clouds \cite{zhao2021pointtransformer, guo2021pct, yu2022pointbert} in addition to text. This is achieved through the incorporation of additional domain-specific encoders and decoders.

Recent studies \cite{flamshepherd2022lmmoldist,edwards2022molt5,pei2023biot5,christofidellis2023unifying,nach0} have demonstrated the ability of LMs in processing specialized chemical languages, such as SMILES \cite{weininger1988smiles} and SELFIES \cite{krenn2020selfies}. These models exhibit proficiency in understanding and manipulating textual representations of chemical data, enabling their application across various tasks. For instance, LMs have been utilized for molecular property prediction \cite{ross2022molformer}, molecular generation \cite{edwards2022molt5}, and chemical reaction prediction \cite{irwin2022chemformer, lu2022t5chem}. 

\begin{figure*}
\centering
\begin{subfigure}{.21\textwidth}
  \centering
  \includegraphics[width=.80\linewidth]{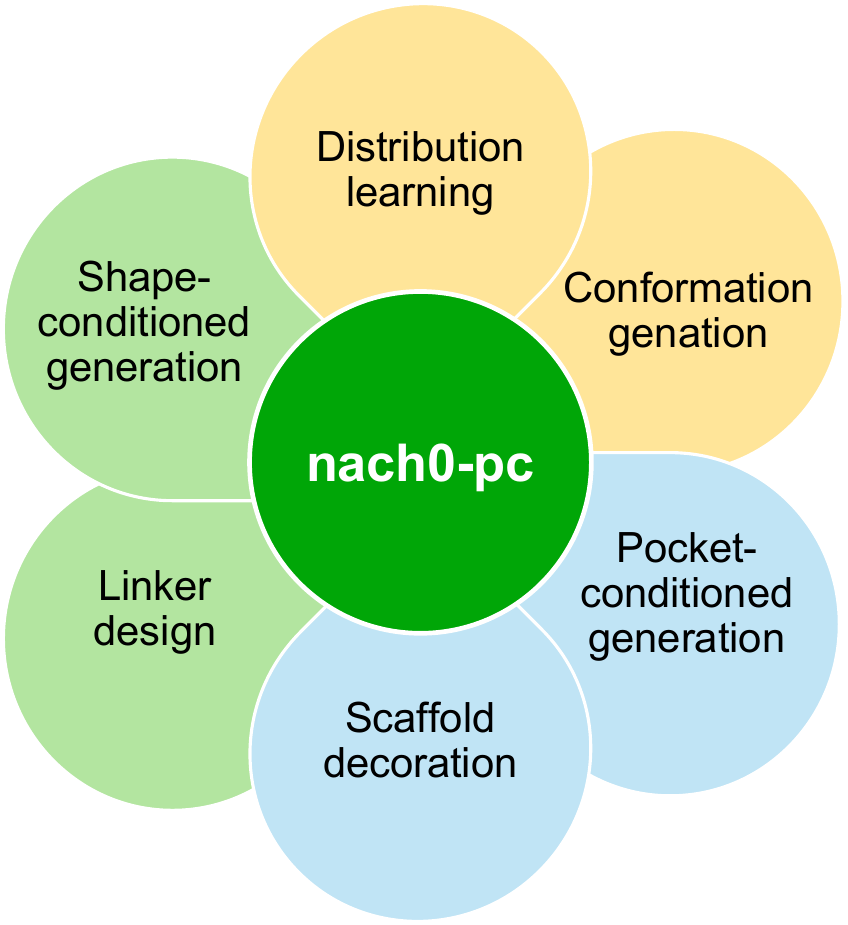}
  \caption{Summary of tasks}
  \label{fig:task}
\end{subfigure}%
\begin{subfigure}{.8\textwidth}
  \centering
  \includegraphics[width=.98\linewidth]{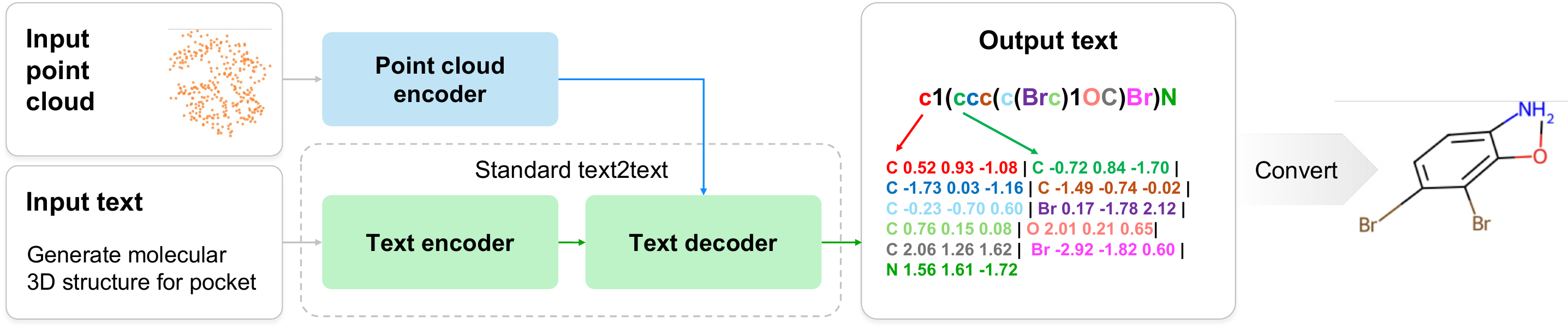}
  \caption{Overview of \ourmodel}
  \label{fig:arch}
\end{subfigure}
\caption{Two diagrams of tasks and \ourmodel{}. (a) Three types of tasks are considered: text→text tasks in yellow; molecular point cloud + text→text in green; and molecular/protein point cloud + text→text in blue. (b) Every spatial molecular generation task we consider is cast as feeding our model text and/or molecular point cloud as input and training it to generate spatial molecular structures as output text.}
\label{fig:overviews}
\end{figure*}

While SMILES and SELFIES capture chemical graph, they lack spatial features crucial for drug discovery methodologies involving precise atom arrangements and interactions. Recent studies \cite{flamshepherd2023xyz} show LMs can generate meaningful 3D chemical structures in text formats like PDB, CIF, and XYZ, representing atom coordinates and features. However, this approach suffers from excessive length, requiring dozens of tokens per atom, becoming impractical for larger structures like proteins. 
Additionally, these formats lack information on atom connectivity, necessitating the use of external software tools like Open Babel\cite{oboyle2011openbabel} to determine chemical bonds. Such external tools are sensitive to noise in atom positions, which can significantly alter reconstructed chemical graphs or cause molecular fragmentation.

In our study, we introduce \ourmodel{}, a Language Model designed for generative tasks involving 3D molecular structures. This model combines domain-specific encoder and textual representation of spatial atom arrangements, enabling effective handling of 3D molecular structures both as inputs and outputs. We leverage a molecular point cloud encoder to derive a concise, meaningful, and order-invariant representation of molecular and protein 3D structures. Furthermore, we propose a textual format for generating spatial molecular structures by initially generating SMILES representations, followed by specifying atom coordinates in accordance with the SMILES sequence order. This format allows LM determine molecular graph and eliminates the reliance on external software for reconstructing bonds.

The key contributions of our work are as follows:
\begin{enumerate} 
    \item A novel \ourmodel~  model that enhances standard encoder-decoder language models by integrating a specialized molecular point cloud encoder and tokens, incorporating unique features like point embedding calculation and position bias adjustments tailored for point cloud data. 

    \item A new pre-training approach that uses dropout on entire subfragments to train the model to predict missing parts of incomplete molecular point clouds, generated through fragment omitting and/or  blurring strategies.

    \item Extensive experiments demonstrating superior or comparable performance to LM baselines and state-of-the-art diffusion approaches across six spatial molecular generation tasks. 
\end{enumerate}

\section{Related Work}

\paragraph{Language Models in Chemistry}

The sequential nature of molecules enables the use of Transformer models and pre-training methods with models like ChemBERTa \cite{chithrananda2020chemberta}, T5Chem \cite{lu2022t5chem}, ChemFormer \cite{irwin2022chemformer}, and BARTSmiles \cite{chilingaryan2022bartsmiles}, which utilize masked language modeling for molecular SMILES.  Recent advancements have introduced domain-specific LMs \cite{edwards2022molt5,christofidellis2023unifying,pei2023biot5}, based on T5, designed specifically to incorporating both chemical and linguistic knowledge. MolT5 \cite{edwards2022molt5} uses initial pre-training on a collection of molecule SMILES and texts, followed by single-task fine-tuning on downstream tasks. On the other hand, Text+Chem T5 \cite{christofidellis2023unifying} is a cross-domain, multi-task T5 model fine-tuned on five tasks, including forward reaction prediction and retro-synthesis. Another recent model, multi-domain nach0 \cite{nach0}, undergoes fine-tuning on a diverse set of 28 task-dataset pairs, employing instruction tuning in a multi-task fashion.

\paragraph{Spatial molecular structure generative models}
The majority of recently published spatial molecular structure generative models is based on the denoising diffusion probabilistic model (DDPM) paradigm \cite{sohldickstein2015diffusion, ho2020diffusion}. These models employ a sequential denoising process, wherein the initial step involves the allocation of positions sampled from a Gaussian distribution model, followed by iterative elimination of noise to construct the molecular structure. EDM \cite{hoogeboom2022edm} follows this methodology to solve \textbf{spatial molecular distribution learning}. 
A notable drawback of the diffusion approach for generating 3D molecules is its reliance on external software like OpenBabel to reconstruct molecular bonds from atom coordinates, highly sensitive to slight errors in the reconstruction process. This limitation has been addressed in several further works. The MolDiff model \cite{peng2023moldiff} integrates an additional bond predictor to guide diffusion and ensure bond consistency alongside 3D atom coordinates. The MDM model \cite{huang2023mdm} integrates a diffusion model with a SchNet encoder \cite{schutt2017schnet} and a scoring network to ensure edge consistency and enhance sample diversity.
GeoDiff \cite{xu2022geodiff}, Torsional Diffusion \cite{jing2022torsionaldiffusion} models can take molecular graph to perform \textbf{conformation generation} task. GeoMol\cite{ganea2021geomol} is also a significant advancement in the field of cheminformatics and drug discovery. It is an end-to-end, non-autoregressive, and SE(3)-invariant machine learning approach to generate distributions of low-energy molecular 3D conformers. DiffLinker \cite{igashov2022difflinker} and LinkerNet \cite{guan2023linkernet} models are a 3D equivariant diffusion model learned to generates the \textit{linker} fragment between given disconnected molecular subfragments in \textbf{linker design} task.  These models can find a stable linker and connect the fragments, resulting in a low-energy conformation for the entire molecule. DiffDec \cite{xie2024diffdec} model was proposed for \textbf{scaffold decoration} task and generates R-groups given a core of the molecule called \textit{scaffold}.
The ShapeMol \cite{chen2023shapemol} model utilizes a diffusion approach for \textbf{shape-conditioned generation}, where a reference ligand molecule is represented as a blurred spatial area approximating the molecular surface volume, and the model suggests new molecules with shapes closely resembling the reference.
The \textbf{pocket-conditioned generation} task involves creating molecules that seamlessly fit within a designated pocket space, establishing favorable interactions to enhance binding affinity. Diffusion models like D3FG\cite{lin2023d3fg}, TargetDiff\cite{guan2023targetdiff}, and DecompDiff \cite{guan2023targetdiff} are capable of designing novel 3D molecules from scratch that effectively bind to specified protein pockets, with DecompDiff using data-dependent decomposed priors reflecting ligand segmentation into functional regions.

Additional related work is presented in Supplementary.
\section{Architecture}

As shown in Fig. \ref{fig:arch}, the architecture of the proposed \ourmodel{} model extends the standard encoder-decoder LM with a domain-specific molecular point cloud encoder and point cloud tokens in a plug-and-play manner. We use a base encoder-decoder LM; specifically, the T5 architecture \cite{raffel2020t5}.

The method's plug-and-play nature allows one to choose whether to train a model from scratch or fine-tune a pre-trained LM model alongside a point cloud encoder for text and point cloud tasks. We initialize \ourmodel’s LM component with nach0 \cite{nach0}, the state-of-the-art natural language and chemical LM. Ablation of LM components is presented in Supplementary material.

\subsection{Input/Output Data Format}

The model accepts textual input and \textit{optionally} molecular point cloud input, while the output is textual.
An input molecular point cloud consists of an unordered collection of points, denoted as $\{ p_i \}$, where each point $p_i = (c_i, f_i)$ is described by Cartesian coordinates $c_i=(x_i,y_i,z_i)$ and an unordered set of tokens $f_i=\{ f_i^j \}$ representing its features. For example, the features of a point corresponding to a ligand's atom might include its atom symbol and atom charge, such as $f_i =$ \texttt{\{‘ligand’, ‘N’, ‘+’\}}. The feature set for a protein's atom point can be extended to encompass its atom name and the amino acid name it belongs, such as $f_i =$ \texttt{\{‘pocket’, ‘C’, ‘GLY’, ‘CA’\}}. All point features are represented as word tokens and can be utilized in textual input and output. For instance, atom symbols and charges are present in point features and SMILES tokens. 

We represent a 3D molecule as text by combining SMILES and XYZ formats. Starting with SMILES to describe the molecular graph, we concatenate lines of XYZ format, describing each atom's symbol and positions in the same order they appear in SMILES. This direct fusion of formats eliminates the need for external software to reconstruct the molecular graph. To limit token count, we use two digits after the decimal point and tokenize each coordinate by splitting at the decimal point (\texttt{‘-1.23’} \textrightarrow \texttt{[‘-1’,‘.23’]}), thus each coordinate can be described by two tokens. Totally it takes 12 tokens in average to describe one atom.

In our work we consider hydrogen-depleted molecular graphs for input point cloud and output SMILES+XYZ representation. To prepare the input point clouds, we center coordinates by deducting the average position from each point. To augment inputs and outputs, we apply the random rotations to the point clouds and make use of non-canonical SMILES representations for textual inputs/outputs.

For large ligand and pocket point clouds, we perform prioritized point downsampling - we keep ligand, C-alpha, and terminal atoms of protein amino acids and downsample other points. This procedure reduces the number of points to the fixed number and makes the model memory efficient for larger point clouds.

\subsection{Point Cloud Encoder}

\begin{figure}[t]
\begin{center}
\centerline{\includegraphics[width=\columnwidth]{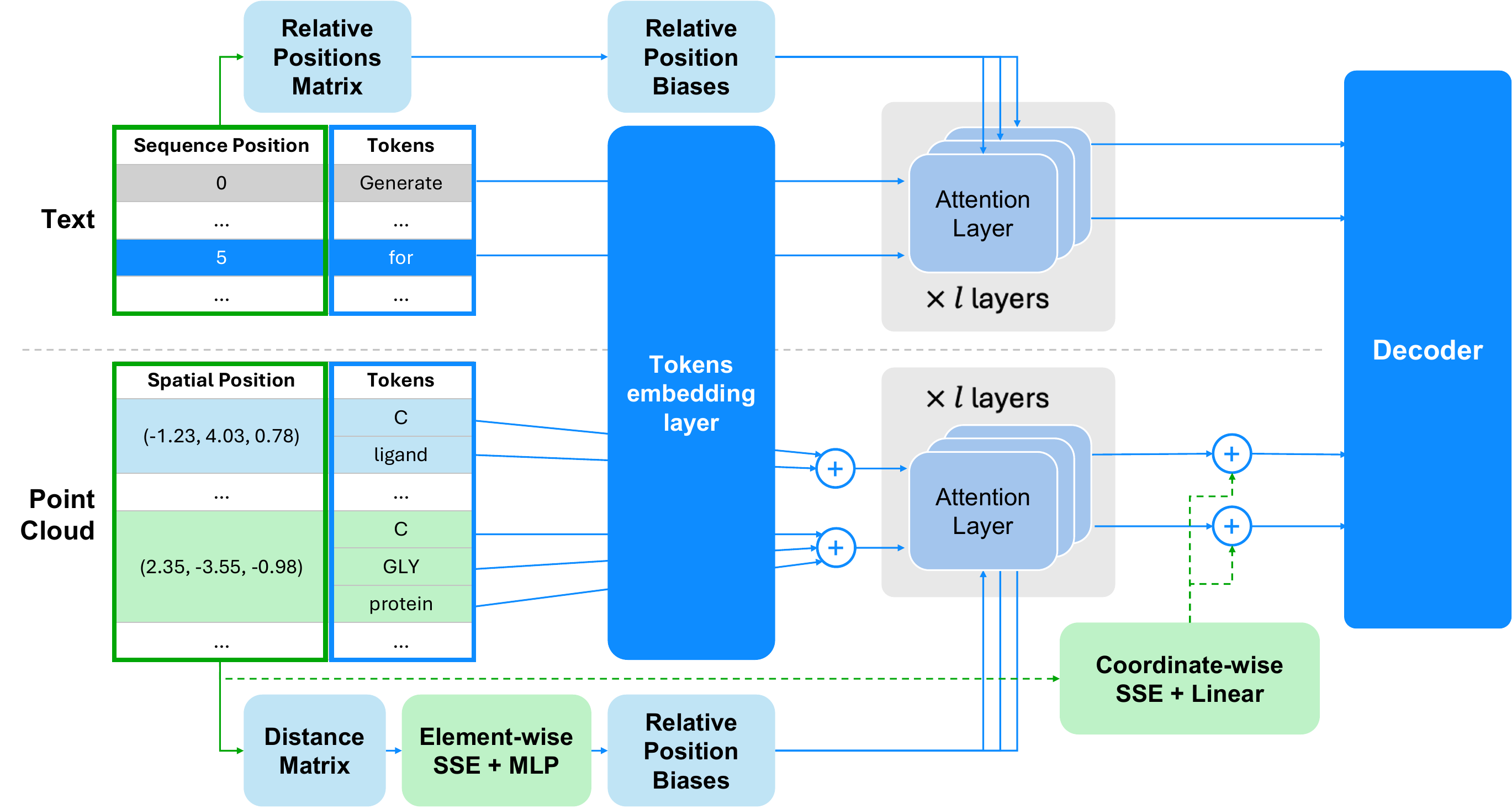}}
\caption{An overview of the encoder architecture adapted for point cloud data and standard text input. For point clouds, tokens represent features at specific spatial positions. Tokens are embedded via a token embedding layer, followed by summation pooling to optimize memory and processing efficiency. Scalar Sinusoidal Embeddings (SSE) integrate continuous spatial coordinates and relative pairwise distances.}
\label{fig:encoders}
\end{center}
\end{figure}

The point cloud encoder utilizes standard language models text encoder architecture with several changes (see Fig. \ref{fig:encoders}) inspired by the nature of point cloud data.  Unlike traditional text, the fundamental data unit is a point rather than a word token. Each point has features and a 3D position.

The first key difference is the point embedding calculation. While word tokens are transformed using a token embedding layer, the same layer is applied to point tokens, followed by a summation pooling operation to aggregate point token embeddings. Representing spatial molecular data with point clouds uses far fewer embeddings than conventional text representation.
For example, a molecular point cloud employs only a few dozens of embeddings whereas text-based representations require several hundreds of embeddings.
It significantly decreases memory usage while increasing the speed of attention layers sensitive to input size.

\begin{table*}[t]
\begin{center}
\begin{small}
\begin{tabular}{ | c | c || c | c | c || c | c |} 
 \hline
 \multirow{2}{*}{Group} & \multirow{2}{*}{Metrics} & \multicolumn{3}{c||}{State-of-the-art} & \multicolumn{2}{c|}{\ourmodel{}} \\ 
 \cline{3-7}
  & & MolDiff  & OpenLLaMA  & nach0 & Multi-Task & Single-Task \\
 \hline
 \multirow{4}{*}{Basic} & Validity and Connectivity ($\uparrow$) & \textbf{99.3\%} & 95.0\% & 98.6\% & 97.8\% & \underline{99.0\%} \\
  & Uniqueness ($\uparrow$) & 92.8\% & \textbf{99.8\%} & \textbf{99.8\%} & \underline{99.2\%} & \textbf{99.8\%} \\
 & Novelty ($\uparrow$) & \underline{97.7\%} & 85.0\% & \textbf{98.0\%} & 96.2\% & 97.1\% \\
 & Diversity ($\uparrow$) & \underline{0.763} & 0.745 & 0.739 & \textbf{0.781} & 0.734 \\
 \hline
 \multirow{3}{*}{Druglikeness} & QED ($\uparrow$) & \underline{0.679} & 0.673 & 0.667 & \textbf{0.771} & 0.664 \\
  & SA ($\uparrow$) & \textbf{0.875} & 0.808 & 0.851 & \underline{0.872} & 0.848 \\ 
  & Lipinski ($\uparrow$) & \underline{4.981} & 4.972 & 4.938 & \textbf{4.992} & 4.938 \\
 \hline
\multirow{3}{*}{3D substructures} & JS. bond lengths ($\downarrow$) & 0.436 & 0.257 & \underline{0.148} & 0.193 & \textbf{0.142} \\
 & JS. bond angles ($\downarrow$) & 0.181 & 0.136 & \underline{0.096} & 0.100 & \textbf{0.94} \\ 
 & JS. dihedral angles ($\downarrow$) & 0.198 & \textbf{0.110} & \textbf{0.110} & 0.132 & \underline{0.113} \\
 \hline 
\multirow{4}{*}{Bonds} & JS. \# bonds per atoms ($\downarrow$)  & 0.121 & \textbf{0.064} & 0.111 & 0.233 & \underline{0.094}  \\
 & JS. freq. bond types  ($\downarrow$) & 0.170 & \textbf{0.031} & 0.046 & 0.052 & \underline{0.041} \\
 & JS. freq. bond pairs  ($\downarrow$) & 0.153 & \textbf{0.028} & 0.037 & 0.040 & \underline{0.033} \\
 & JS. freq. bond triplets  ($\downarrow$) & 0.137 & \textbf{0.034} & 0.046 & 0.054 & \underline{0.041} \\
\hline
\multirow{3}{*}{Rings} & JS. \# rings ($\downarrow$) & 0.079 & \textbf{0.033} & 0.56 & 0.270 & \underline{0.036} \\
 & JS. \# n-sized rings ($\downarrow$) & 0.102 & \textbf{0.024} & \underline{0.026} & 0.058 & \textbf{0.024} \\
 & \# Intersecting rings ($\uparrow$) & \underline{8} & \underline{8} & \underline{8} & \textbf{9} & \underline{8} \\  
\hline
\end{tabular}
\end{small}
\end{center}
\caption{Spatial molecular distribution learning performance metrics on GEOM-DRUGS.}
\label{tab:distribution_learning_3d}
\end{table*}

The second notable distinction can be found in point spatial coordinates.
Contrary to textual representation, which is a sequence of discrete token indices, every point's position in a point cloud is depicted using continuous Cartesian coordinates.
First, we modify the relative position biases computation to embed pairwise distances instead of relative sequence positions. Also, we embed the coordinates of the point and sum them with point embeddings. We do it on the last step right before passing embeddings to the decoder, ensuring that the self-attention layers remain invariant to any point cloud translation or rotation.

To embed scalar continuous values of distances and coordinates, we are introducing Scalar Sinusoidal Embeddings (SSE). It takes inspiration from positional sinusoidal embeddings \cite{vaswani2017transformer} and extends it to map continuous scalar values into a high-dimensional vector. We divide the input scalar $s$ by wavelengths $w_i$, uniformly initialized on a logarithmic grid, and use the yielded sine and cosine function values to form the resulting embedding vector. 

$$ SSE_{2i}(s) = sin(s / w_i), SSE_{2i+1}(s) = cos(s / w_i) $$

The step-by-step description can be found in Alg. \ref{alg:pcencoding}.

\begin{algorithm}[h!]
\caption{Point Cloud Encoder}
\label{alg:pcencoding}
\begin{algorithmic}[1]
  \Require Point features $f_i=\{ f_i^j \}$ and pos $c_i=(x_i,y_i,z_i)$
  \Ensure Point embeddings $n^{out}_{i}$
  \State $n^{0}_{i} = \sum_{f_i^j \in f_i} Emb(f_i^j)$ \Comment{embed and aggregate feats}
  \For{$l = 1, 2, \dots, L$}  
    \State $b^{l}_{hij} = MLP^{l}_{h}(SSE(||c_i - c_j||))$ \\ \Comment{for each head $h$ compute relative attention biases}
    \State $n^{l} = SelfAttention^{l}(n^{l-1}, b^{l})$ \Comment{update  embs}
  \EndFor
  \State $n^{out}_{i}=n^{L}_{i}+Lin_{xyz}(SSE(x_i), SSE(y_i), SSE(z_i)) $ \\ \Comment{embed coordinates}
\end{algorithmic}
\end{algorithm}

\subsection{Molecular Point Cloud Pre-training}

\begin{figure}[t]
\begin{center}
\centerline{\includegraphics[width=\columnwidth]{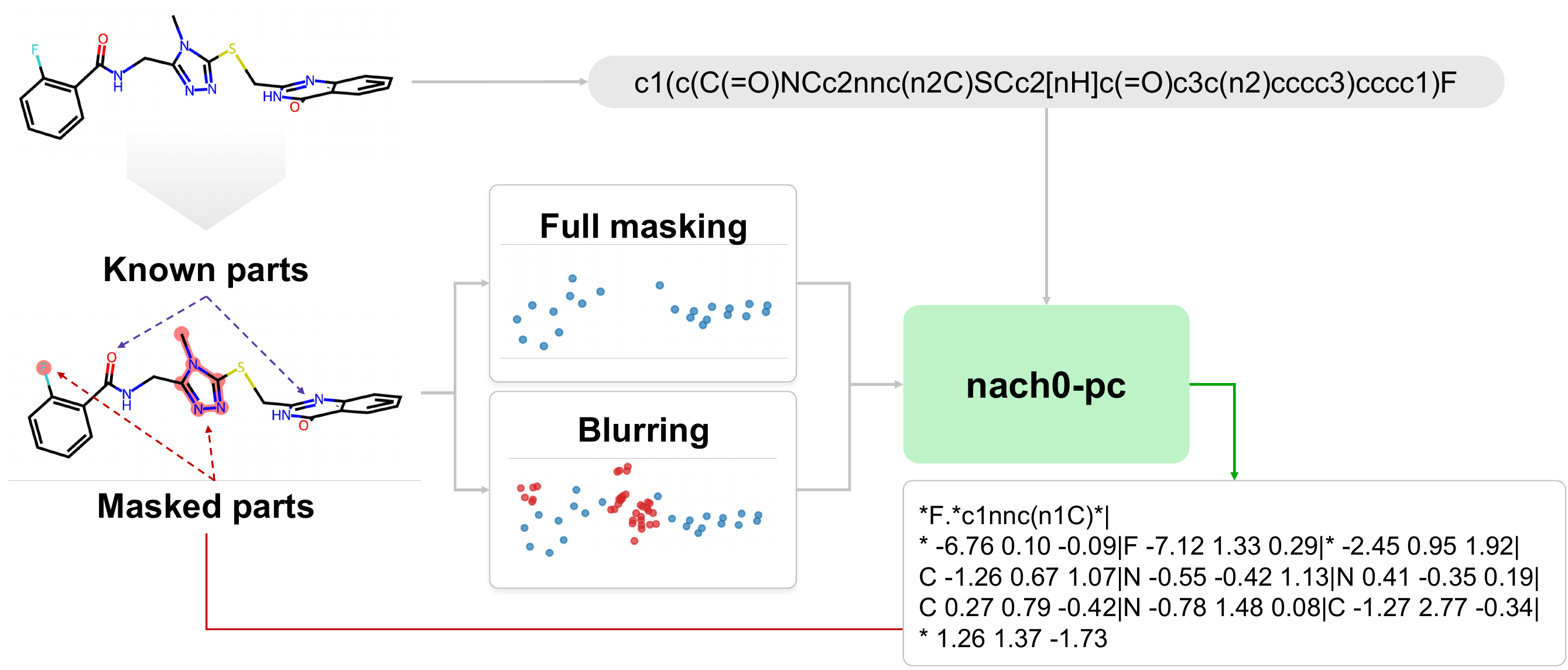}}
\caption{The pre-training scheme for 3D molecular structures datasets. The model learns to reconstruct blurred or masked molecular fragments. The model generates the missing parts during pre-training, including their SMILES representations, attachment points, and atom coordinates.}
\label{fig:pretraining}
\end{center}
\end{figure}

Drawing inspiration from the token dropout pre-training objective of T5 \cite{raffel2020t5}, we introduce a novel pre-training scheme for molecular point clouds to distillate the knowledge from unlabeled spatial molecular structures datasets (Fig. \ref{fig:pretraining}). In our approach, we feed the model with incomplete molecular point clouds and train the model to reconstruct the missing pieces. It is important to mention that our dropout method operates with whole subfragments rather than individual atoms. We employ the BRICS \cite{degen2008brics} algorithm to split the molecule into several fragments and randomly select a subset of these fragments with some predefined probability (ensuring that at least one is chosen) and exclude them from the molecule. 
The pre-training scheme is designed to be flexible, handling both datasets consisting only of spatial molecular structures and datasets with protein pockets and ligand pairs. In the latter, the ligand is masked, while the protein pocket remains unchanged in the input point cloud.

Further, we form the input point cloud by (i) removing or (ii) obfuscating chosen fragments by omitting the features of a point (such as atomic symbol, charge, and valency), replicating the point multiple times, and introducing Gaussian noise to the coordinates of these copies. 

During pre-training, the model's task is to accurately recover the absent (or blurred) components, specifying their SMILES representations as well as their atomic coordinates. Each recovered fragment should include a connecting point indicated by the symbol \texttt{‘*’}. In cases where the model should reconstruct multiple missing fragments, the fragments are separated with the \texttt{‘.’} token. We note that 3D pre-training enhances the model's performance on downstream tasks (see analysis of the 3D pre-training impact in Supplementary material).

\begin{figure*}[t]
\centering
\begin{subfigure}{.25\textwidth}
  \centering
  \includegraphics[width=.95\linewidth]{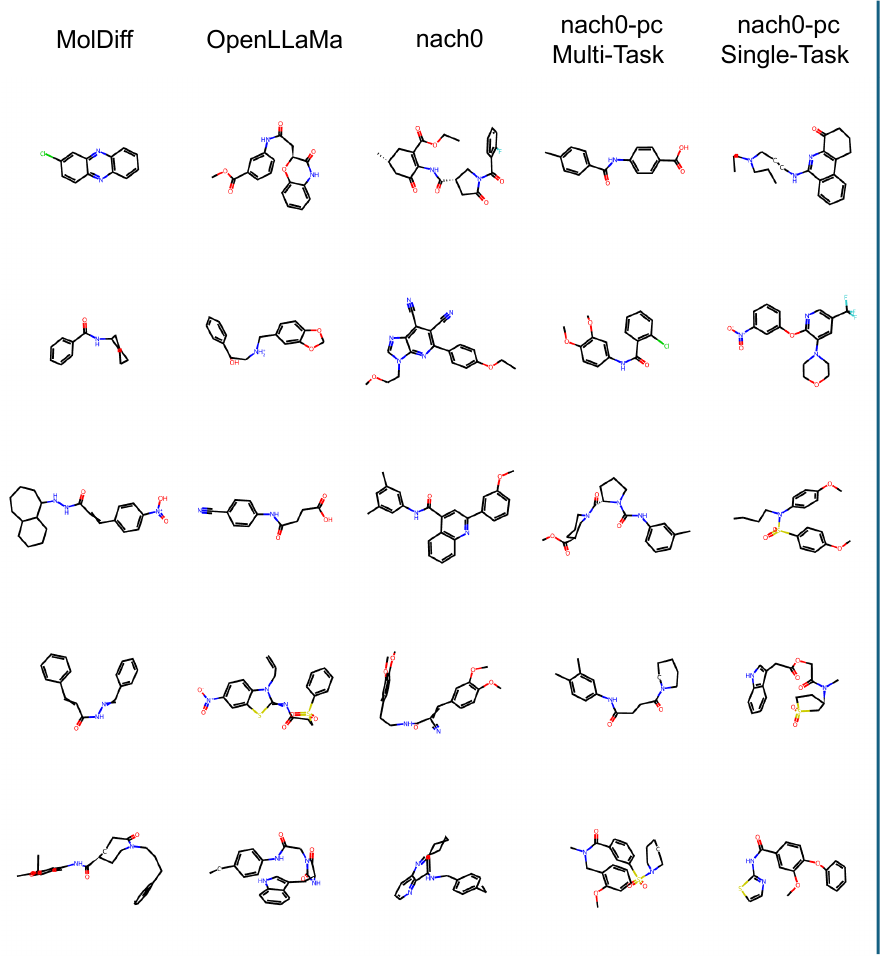}
  \caption{Distribution learning}
  \label{fig:distr_learn_samples}
\end{subfigure}%
\begin{subfigure}{.35\textwidth}
  \centering
  \includegraphics[width=.95\linewidth]{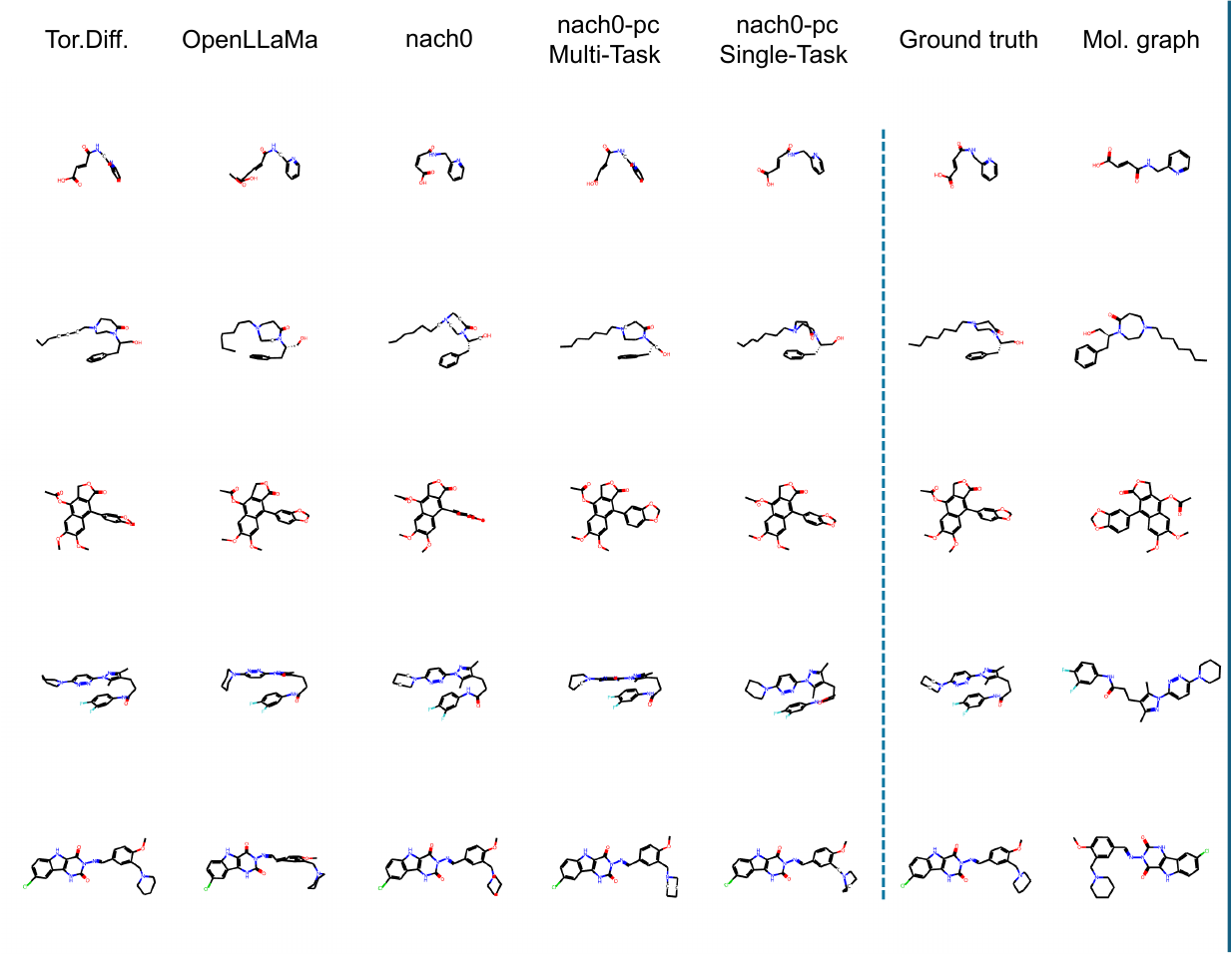}
  \caption{Conformation generation}
  \label{fig:conf_gen_samples}
\end{subfigure}
\begin{subfigure}{.30\textwidth}
  \centering
  \includegraphics[width=.95\linewidth]{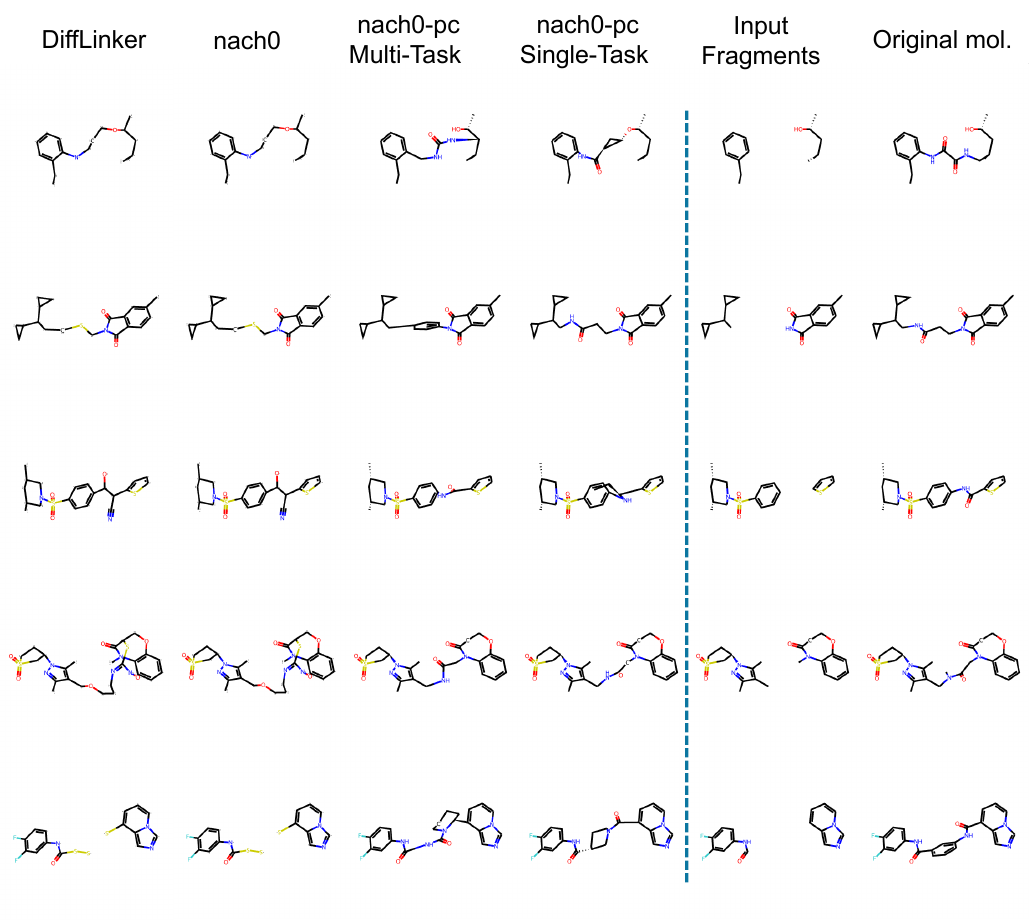}
  \caption{Linker design}
  \label{fig:linker_design}
\end{subfigure}%
\caption{Samples for (a) spatial distribution learning, (b) conformation generation and (c) linker design tasks.}
\label{fig:distr_learn_conf_gen_samples}
\end{figure*}
\section{Experiments}

We evaluate the quality of the \ourmodel{} model across several established spatial molecular generation tasks: (i) 3D molecular structures generation: spatial molecular distribution learning, conformation generation, (ii) molecular  completion: linker design, scaffold decoration, (iii) shape-conditioned generation, (iv) pocket-conditioned generation. We highlight \textbf{the best} and \underline{the second best} metrics values for better readability and provide sampled spatial molecular structures for visual inspection of results.

We compare our \ourmodel{} model with well-established baseline models proposed for each mentioned task. We also compare it with text-only language model - we train decoder-only OpenLLaMA model \cite{openlm2023openllama} with the same (350M) number of parameters as \ourmodel{}, along with nach0 \cite{nach0} on spatial molecular distribution learning, conformation generation and linker design tasks, where input/output text and molecular condition in SMILES+XYZ format can be described with fewer than a thousand tokens. The rest of tasks either require large protein pocket condition 
or  involve blurring that substantially inflate the size of the point cloud representation, making these tasks practically unfeasible in text-only paradigm.

We train our proposed \ourmodel{} model in both multi-task and single-task regimes to fairly evaluate it against both LM and non-LM models. Our work adopts small molecules ZINC \cite{zinc20-dataset}, MOSES \cite{polykovskiy2020moses}, and GEOM-Drugs \cite{axelrod2022geom} datasets, as well as the CrossDocked2020 \cite{francoeur2020crossdocked} dataset, which includes pocket-ligand pairs. In cases when tasks use the same dataset, to avoid any potential data leakage, we use the same dataset split. We combine all stated datasets to both pretrain and finetune the \ourmodel{} model in multi-task regime. To address the discrepancy in dataset sizes, we normalize the training process by employing a batch-balancing technique that involves retrieving random object from uniformly sampled task. 

See Supplementary material for additional details on datasets, models, and metrics.

\subsection{Spatial Molecular Distribution Learning and Conformation Generation Tasks}
\label{sec:texttotext_tasks}

This section evaluates the model's ability to generate structurally and physically plausible spatial molecular objects. 

The \textbf{spatial molecular distribution learning} task assesses whether the model can produce novel 3D molecular structures whose distribution is close to the ground truth. Following \cite{peng2023moldiff}, we adopt high-quality GEOM-Drugs \cite{axelrod2022geom} dataset, which offers gold-standard conformation ensembles generated using metadynamics in CREST \cite{pracht2020automated}. We evaluate generated molecules from several perspectives, including basic (validity, novelty, etc)  drug-likeness, and 3D substructures, bonds and rings distribution divergences.

\begin{table}
\begin{center}
\begin{small}
\begin{tabular}{| c | c | c | c | c | c | c | c | c | } 
 \hline
\multirow{3}{*}{Method} & \multicolumn{4}{c|}{Recall} & \multicolumn{4}{c|}{Precision} \\
\cline{2-9} 
  & \multicolumn{2}{c|}{COV ($\uparrow$)} & \multicolumn{2}{c|}{AMR ($\downarrow$)} & \multicolumn{2}{c|}{COV ($\uparrow$)} & \multicolumn{2}{c|}{AMR ($\downarrow$)} \\ 
 \cline{2-9} 
 & Avg & Med & Avg & Med & Avg & Med & Avg & Med  \\
 \hline
 \multicolumn{9}{|c|}{Single-Task baselines} \\
 \hline
 ETKDG  & 38.4 & 28.6 & 1.06 & 1.00 & 40.9 & 30.8 & 0.99 & 0.89 \\
 GeoMol & 44.6 & 41.4 & 0.87 & 0.83 & \underline{43.0} & \underline{36.4} & \underline{0.93} & \underline{0.84} \\
 GeoDiff & 42.1 & 37.8 & 0.83 & 0.81 & 24.9 & 14.5 & 1.14 & 1.09 \\
 Tor. Diff.  & \textbf{72.7} & \textbf{80.0} & \textbf{0.58} & \textbf{0.56} & \textbf{55.2} & \textbf{56.9} & \textbf{0.78} & \textbf{0.73} \\
 \hline
 \multicolumn{9}{|c|}{Multi-Task Text-Only LM baselines} \\
 \hline
 OpenLLaMA & 10.2 & 0.8 & 1.48 & 1.43 & 19.7 & 1.7 & 1.31 & 1.28 \\ 
 nach0 & 54.0 & 54.55 &  0.74 & 0.73 & 30.7 & 20.7 & 1.06 & 1.06 \\  
 \hline
 \multicolumn{9}{|c|}{\ourmodel{}} \\
 \hline
 Multi-Task & 50.1 & 50.0 & 0.76 & 0.75 & 27.7 & 16.7 & 1.11 & 1.11 \\
 Single-Task & \underline{57.7} & \underline{59.5} & \underline{0.70} & \underline{0.69} & 32.4 & 23.1 & 1.03 & 1.03 \\
 \hline
\end{tabular}
\end{small}
\end{center}
\caption{Generated conformer ensembles quality on GEOM-DRUGS (baselines' metrics from \cite{jing2022torsionaldiffusion}).}
\label{tab:conf_gen}
\end{table}

\begin{figure*}[t]
\centering
\begin{subfigure}{.55\textwidth}
  \centering
  \includegraphics[width=.95\linewidth]{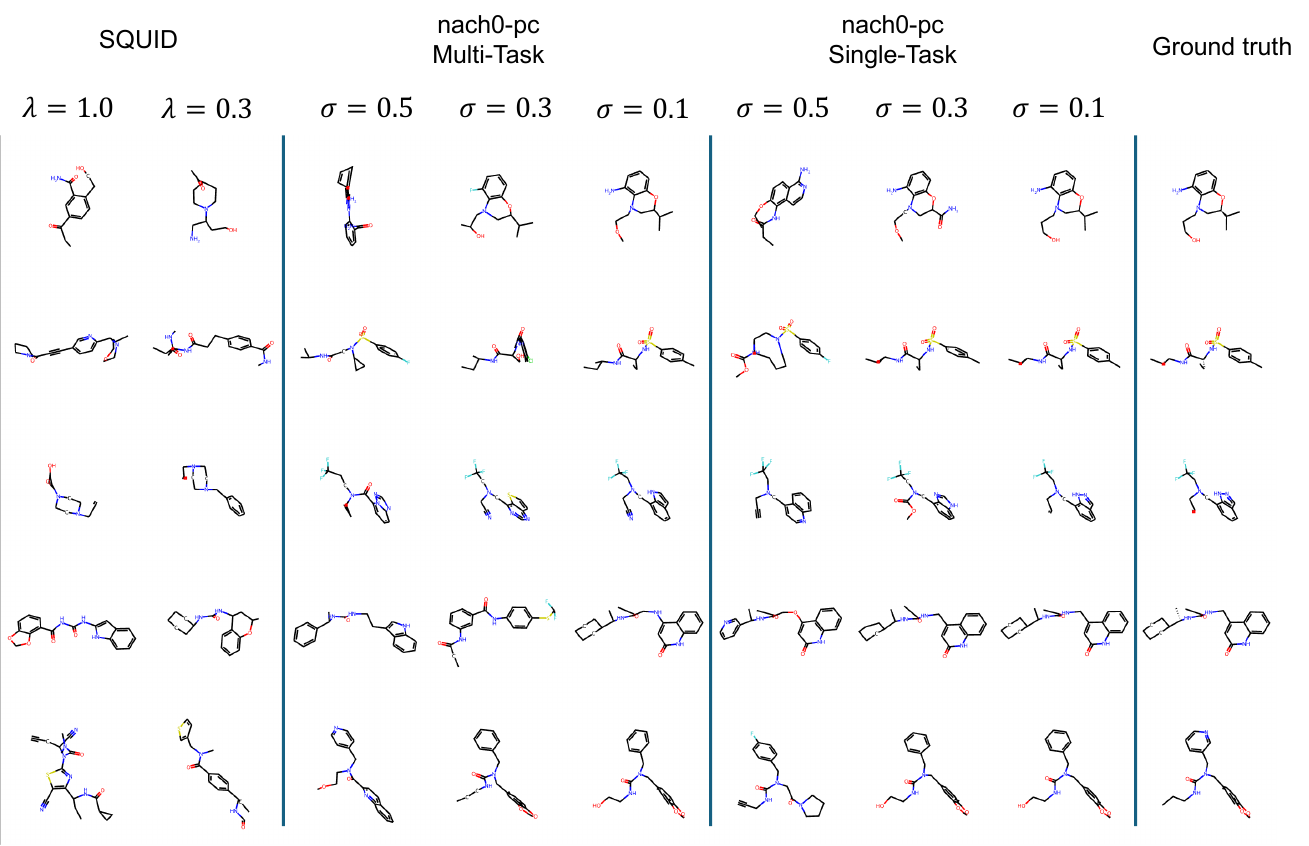}
\end{subfigure}%
\begin{subfigure}{.45\textwidth}
  \centering
  \includegraphics[width=.95\linewidth]{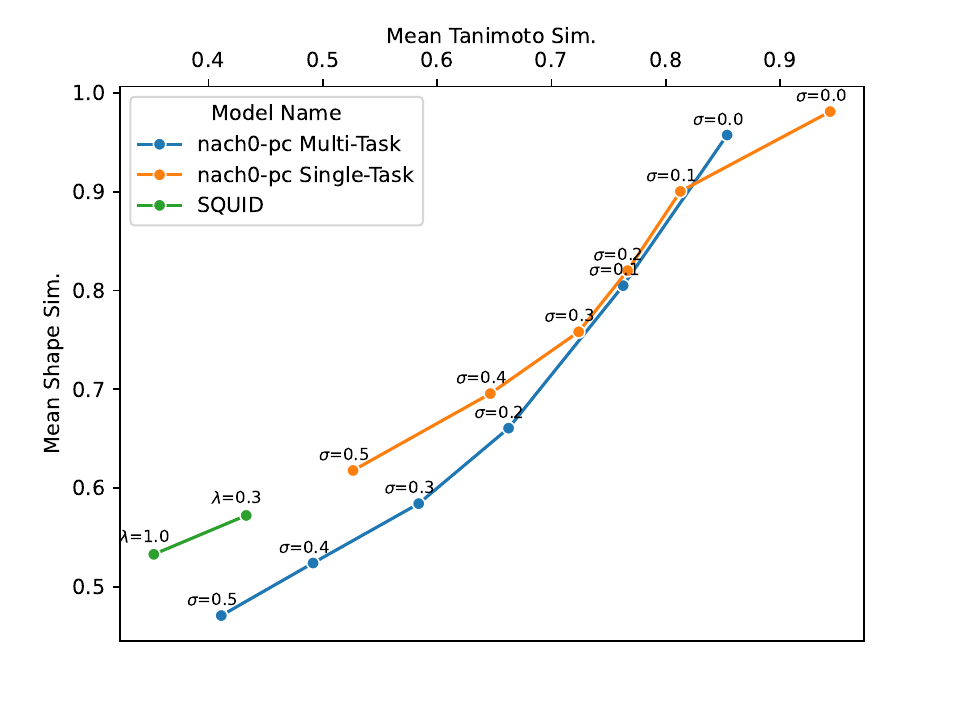}
\end{subfigure}
\caption{(left) Generated molecular structures and (right) structural/shape similarity trade-off for various noise parameters on shape-conditioned generation task.}
\label{fig:shape_conditioned}
\end{figure*}

In the \textbf{conformation generation}, we focus on generating plausible \textit{conformations} given molecular graph. The \textit{conformations} of a molecule are its energetically favorable 3D structures, each representing a local minimum on the potential energy surface. Similar to the previous task, we employ the GEOM \cite{axelrod2022geom} dataset for this task. We follow \cite{jing2022torsionaldiffusion} and employ the Average Minimum RMSD (AMR) and Coverage metrics. These metrics are evaluated from Recall and Precision view. 

We utilize the same train/validation/test splits as in the conformation generation task from the Torsional Diffusion\cite{jing2022torsionaldiffusion} paper and retrain baseline if they were trained on another split.

The results for two tasks are presented in Tables \ref{tab:distribution_learning_3d} and \ref{tab:conf_gen}. Also, one can find the examples of generated structures on Figs. \ref{fig:distr_learn_samples}, \ref{fig:conf_gen_samples}. The proposed \ourmodel{} model shows the best or second best results for the majority of metrics on both first and second tasks. As shown in Tab. \ref{tab:distribution_learning_3d}, text-only OpenLLaMA shows slightly better results on bonds and rings distribution metrics. Our hypothesis, it is due to the fact the input of this task is the same all time, so decoder-only model better utilize parameters rather than encoder-decoder where the parameters are divided between model parts. On the conformation generation task, the only model outperforming \ourmodel{} on the whole set of metrics is Torsional Diffusion, which utilizes samples from external software RDKit \cite{landrum2023rdkit} as an initial generation point on the inference stage. Nevertheless, \ourmodel{} shows stronger results than all other purely neural baselines that produce molecular conformation from scratch.

\begin{table}
\begin{center}
\begin{small}
\begin{tabular}{| c | c | c | c | c | c |} 
 \hline
 Method & Val. ($\uparrow$) & Uniq. ($\uparrow$) & Filt. ($\uparrow$) & RMSD ($\downarrow$) & $SC$ ($\uparrow$) \\
 \hline
 \multicolumn{6}{|c|}{State-of-the-art} \\
 \hline
 DeLinker & \textbf{98.3\%} & \textbf{44.2\%} & 84.88\% & 5.48 & 0.49 \\
 3DLinker & 71.5\% & \underline{29.2\%} & 83.72\% &  \textbf{0.11} & \underline{0.92} \\
 DiffLinker & \underline{93.8\%} & 24.0\% & 86.26\%  & \underline{0.34} & \textbf{0.93} \\
 nach0 & 55.2\% & 19.9\% & 98.78\% & 1.41 & 0.85 \\
 \hline
 \multicolumn{6}{|c|}{\ourmodel{}} \\
 \hline
 Multi-Task & 81.6\% & 27.6\% & \underline{99.00\%} &  1.28 & 0.86 \\
 Single-Task & 89.7\% & 12.3\% & \textbf{99.55\%}  & 1.04 & 0.88 \\
 \hline
\end{tabular}
\end{small}
\end{center}
\caption{Models performance evaluation for linker design task (metrics for baselines from \cite{igashov2022difflinker}).}
\label{tab:linker_design}
\end{table}

\begin{table}
\begin{center}
\begin{small}
\begin{tabular}{| c | c | c | c |} 
 \hline
 Model & Validity ($\uparrow$) & Uniqueness ($\uparrow$) & QVina Score ($\downarrow$)\\
 \hline
 Reference & - & -  & -8.47 \\
 Scaffolds & - & -  & -7.73 \\
 \hline
 \multicolumn{4}{|c|}{State-of-the-art} \\
 \hline
 Pocket2Mol  & 51.14\% & 44.27\% & -8.11 \\
 FLAG  & 87.95\% & \textbf{65.30}\% & -7.62 \\
 DiffDec  & 98.00\% & \underline{48.54\%} & \textbf{-8.25} \\
 \hline
 \multicolumn{4}{|c|}{\ourmodel{}} \\
 \hline 
 Multi-Task & \underline{97.63\%} & 42.58\% & -7.931 \\
 Single-Task & \textbf{99.17}\% & 18.12\% & \underline{-8.123} \\
 \hline
\end{tabular}
\end{small}
\end{center}
\caption{Scaffold decoration task metrics (metrics of baselines from \cite{xie2024diffdec}).}
\label{tab:scaffold_decoration}
\end{table}

\subsection{Shape-conditioned Generation} \label{sec:shape_cond}

This section focuses on producing molecules spatially similar to the reference structure but structurally dissimilar. This is achieved by representing the reference molecule as a \textit{shape} - an area where molecular atom nucleus and electron clouds are located.  We represent molecular shape as a point cloud - we replicate each atom several times and add Gaussian noise with predefined standard deviation $\sigma$ to atom positions while removing all point features completely. One can balance between spatial and chemical similarity by alternating the parameter $\sigma$.

Following the methodology outlined in the SQUID paper \cite{adams2023squid}, we conduct training and test-stage sampling using the RDKit \cite{landrum2023rdkit} conformations computed for the MOSES dataset \cite{polykovskiy2020moses}. We provide a comparison of \ourmodel{} with the SQUID \cite{adams2023squid} model in Fig. \ref{fig:shape_conditioned}. We alternate noise injection parameters, standard deviation $\sigma$ for \ourmodel{} and prior interpolation coefficient $\lambda$ for SQUID, to show available trade-offs between structural and shape similarity. \ourmodel{} provides a wider range of available trade-offs, covering a high structural to high shape similarity area. Moreover, \ourmodel{} produces more spatially similar objects for low structural similarity values than SQUID.

\subsection{Linker Generation and Scaffold Decoration Tasks} \label{sec:molecular_completion}

These tasks assess the ability of models to complete disjoint or partially-defined molecular structures.

In \textbf{linker design} task, models operate with several disconnected fragments and should produce small molecular structures that spatially and chemically connect the given fragments and complete into one chemical structure. The same as in the DiffLinker work \cite{igashov2022difflinker}, we employ a subset of the ZINC dataset, comprising $250000$ random molecules with conformations generated using RDKit \cite{landrum2023rdkit}. This dataset also provides a split into input fragments and a linker for each molecule.

In the second task, \textbf{scaffold decoration}, the model takes the core part of the molecule called \textit{scaffold} and complete side-chain specific motifs called R-groups. Usually, scaffold decoration is employed to enhance some molecular properties, for instance, binding affinity with a specific protein. Following DiffDec \cite{xie2024diffdec}, we adopt Multi R-Group Decoration Task on CrossDocked \cite{francoeur2020crossdocked} dataset, containing $100K$ ligand-protein pairs, where each ligand is split into a scaffold and R-groups.

Our \ourmodel{} takes molecular fragments/scaffold and protein pocket, if available, as an input point cloud and produces the linker/R-groups without repeating input atoms. The produced molecular substructures contain the attachment points described by a symbol \texttt{`*'} and coordinates. We utilize these attachment points to combine input fragments with generated ones into a coherent molecule. The model can produce several R-groups in the scaffold decoration task by separating them with a symbol \texttt{`.'}.

We compare \ourmodel{} with DeLinker \cite{imrie2020delinker}, 3DLinker \cite{huang223dlinker}, DiffLinker \cite{igashov2022difflinker} on the linker generation task, and benchmark against LibINVENT \cite{fialkova2022libinvent}, FLAG \cite{zhang2023flag}, and DiffDec \cite{xie2024diffdec} for scaffold decoration. 

As shown in Tables \ref{tab:linker_design} and \ref{tab:scaffold_decoration} for both tasks, \ourmodel{} can complete input molecular point clouds with a high success rate, producing molecules (Fig. \ref{fig:linker_design}) that pass 2D filters such as PAINS \cite{baell2010pains}. Despite moderate structural diversity, \ourmodel{} produces spatially diverse molecules. Moreover, it enhances scaffold binding affinity, working on par with other state-of-the-art models in scaffold decoration.

\begin{table}
\begin{center}
\begin{small}
\begin{tabular}{| c | c | c | c | c | c |} 
 \hline
 \multirow{2}{*}{Model} & \multirow{2}{*}{Valid. ($\uparrow$)} & \multirow{2}{*}{Div. ($\uparrow$)} & \multicolumn{2}{c|}{Vina Dock ($\downarrow$)} & \multirow{2}{*}{\makecell{High \\ Affinity ($\uparrow$)}}\\
 \cline{4-5}
& & & Avg & Med &  \\
 \hline
 Reference & 100\% & -  & -7.45 & -7.26 & - \\
 \hline
 \multicolumn{6}{|c|}{State-of-the-art} \\
 \hline
 AR & \underline{92.95\%} & \underline{0.70} & -6.75 & -6.62 & 37.9\%  \\
 Pocket2Mol  & \textbf{98.31}\% & 0.69 & \underline{-7.15} & -6.79 & \underline{48.4\%}  \\
 TargetDiff  & 90.36\% & \textbf{0.72} & \textbf{-7.80} & \textbf{-7.91} & \textbf{58.1}\%\\
 \hline
 \multicolumn{6}{|c|}{\ourmodel{}} \\
 \hline 
 Multi-Task & 91.78\% & 0.32 & -6.52 & \underline{-6.86} & 38.2\% \\
 Single-Task & 89.82\% & 0.40 & -6.50 & -6.62 & 41.1\% \\
 \hline
\end{tabular}
\end{small}

\end{center}
\caption{Pocket-conditioned generation performance metrics (metrics of baselines from \cite{guan2023targetdiff}).}
\label{tab:pocket_conditioned_generation}
\end{table}

\begin{figure}[t]
\begin{center}
\centerline{\includegraphics[width=\columnwidth]{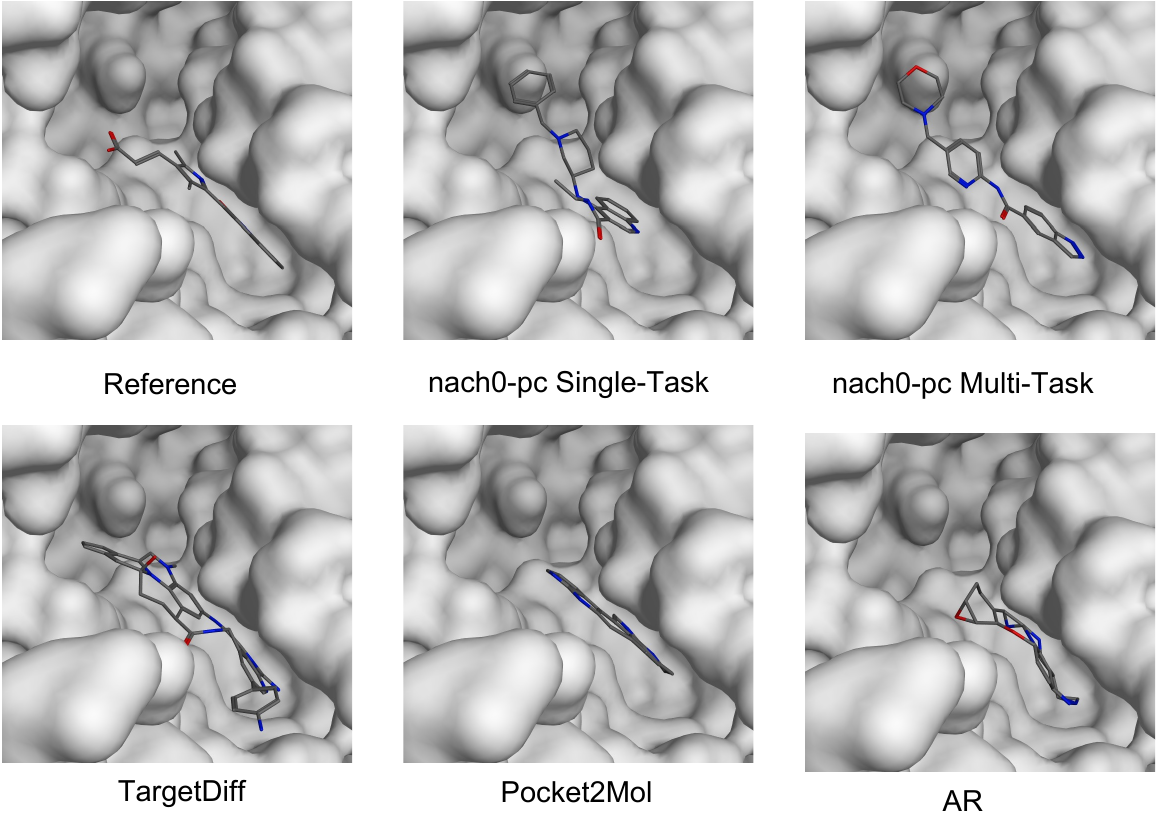}}
\caption{Generated molecular structures for pocket-conditioned generation (reference name \texttt{4iwq\_A}) task.}
\label{fig:pocket_conditioned}
\end{center}
\end{figure}

\subsection{Pocket-conditioned Generation} \label{sec:pocket_cond}

Finally, \ourmodel{} was trained to generate novel high-affinity structures for a given \textit{protein pocket} condition. 

Similar to TargetDiff, we used the CrossDocked2020 dataset \cite{francoeur2020crossdocked} and trained our model on $100000$ high-affinity ligand-protein complexes. 

During the test stage, we randomly sampled $100$ molecules for each protein pocket in the test set. One can find generated ligands visualisation in Fig. \ref{fig:pocket_conditioned}. We assess the validity, diversity and docking scores of generated structures and provide the comparison with AR\cite{luo2021ar}, Pocket2Mol\cite{peng2022pockettomol} and TargetDiff\cite{guan2023targetdiff} baseline models in Tab \ref{tab:pocket_conditioned_generation}. 

Our model shows high validity and diversity. While there is a significant gap between \ourmodel{} and the TargetDiff performance, it demonstrates comparable results to AR and Pocket2Mol based on docking scores and binding affinity.

\section{Conclusion}

We have introduced \ourmodel{}, a novel model adept at generating diverse and physically plausible molecular 3D structures. By combining a domain-specific point cloud encoder with an encoder-decoder language model along with combined SMILES+XYZ textual format and novel molecular point cloud pre-training technique, \ourmodel{} effectively addresses challenges associated with handling chemical 3D structures and a SMILES sequence. Through extensive fine-tuning within single-task and multi-task frameworks, \ourmodel{} exhibits comparable performance to various state-of-the-art diffusion and LM baseline models. As future work, it would be valuable to explore training on a broader range of NLP and Chemistry 2D/3D tasks in a multi-task fashion, including molecular properties prediction based on a spatial input and protein-related tasks. The integration of proposed ideas into decoder-only approaches remains to be explored.

\bibliography{bibliography}

\begin{thebibliography}{75}
\providecommand{\natexlab}[1]{#1}

\bibitem[{Adams and Coley(2023)}]{adams2023squid}
Adams, K.; and Coley, C.~W. 2023.
\newblock Equivariant Shape-Conditioned Generation of 3{D} Molecules for Ligand-Based Drug Design.
\newblock In \emph{The Eleventh International Conference on Learning Representations}.

\bibitem[{Alayrac et~al.(2022)Alayrac, Donahue, Luc, Miech, Barr, Hasson, Lenc, Mensch, Millican, Reynolds, Ring, Rutherford, Cabi, Han, Gong, Samangooei, Monteiro, Menick, Borgeaud, Brock, Nematzadeh, Sharifzadeh, Bi\'{n}kowski, Barreira, Vinyals, Zisserman, and Simonyan}]{alayrac2022flamingo}
Alayrac, J.-B.; Donahue, J.; Luc, P.; Miech, A.; Barr, I.; Hasson, Y.; Lenc, K.; Mensch, A.; Millican, K.; Reynolds, M.; Ring, R.; Rutherford, E.; Cabi, S.; Han, T.; Gong, Z.; Samangooei, S.; Monteiro, M.; Menick, J.~L.; Borgeaud, S.; Brock, A.; Nematzadeh, A.; Sharifzadeh, S.; Bi\'{n}kowski, M.~a.; Barreira, R.; Vinyals, O.; Zisserman, A.; and Simonyan, K. 2022.
\newblock {F}lamingo: a Visual Language Model for Few-Shot Learning.
\newblock In Koyejo, S.; Mohamed, S.; Agarwal, A.; Belgrave, D.; Cho, K.; and Oh, A., eds., \emph{Advances in Neural Information Processing Systems}, volume~35, 23716--23736. Curran Associates, Inc.

\bibitem[{Alhossary et~al.(2015)Alhossary, Handoko, Mu, and Kwoh}]{alhossary2015qvina}
Alhossary, A.; Handoko, S.~D.; Mu, Y.; and Kwoh, C.-K. 2015.
\newblock {Fast, accurate, and reliable molecular docking with {Q}uick{V}ina 2}.
\newblock \emph{Bioinformatics}, 31(13): 2214--2216.

\bibitem[{Axelrod and G{\'o}mez-Bombarelli(2022)}]{axelrod2022geom}
Axelrod, S.; and G{\'o}mez-Bombarelli, R. 2022.
\newblock {GEOM}, energy-annotated molecular conformations for property prediction and molecular generation.
\newblock \emph{Scientific Data}, 9(1): 185.

\bibitem[{Baell and Holloway(2010)}]{baell2010pains}
Baell, J.~B.; and Holloway, G.~A. 2010.
\newblock New Substructure Filters for Removal of Pan Assay Interference Compounds ({PAINS}) from Screening Libraries and for Their Exclusion in Bioassays.
\newblock \emph{Journal of Medicinal Chemistry}, 53(7): 2719--2740.
\newblock PMID: 20131845.

\bibitem[{Bolcato, Heid, and Boström(2022)}]{bolcato2022shapesim}
Bolcato, G.; Heid, E.; and Boström, J. 2022.
\newblock On the Value of Using 3{D} Shape and Electrostatic Similarities in Deep Generative Methods.
\newblock \emph{Journal of Chemical Information and Modeling}, 62(6): 1388--1398.
\newblock PMID: 35271260.

\bibitem[{Brown et~al.(2020)Brown, Mann, Ryder, Subbiah, Kaplan, Dhariwal, Neelakantan, Shyam, Sastry, Askell, Agarwal, Herbert-Voss, Krueger, Henighan, Child, Ramesh, Ziegler, Wu, Winter, Hesse, Chen, Sigler, Litwin, Gray, Chess, Clark, Berner, McCandlish, Radford, Sutskever, and Amodei}]{brown2020gpt}
Brown, T.; Mann, B.; Ryder, N.; Subbiah, M.; Kaplan, J.~D.; Dhariwal, P.; Neelakantan, A.; Shyam, P.; Sastry, G.; Askell, A.; Agarwal, S.; Herbert-Voss, A.; Krueger, G.; Henighan, T.; Child, R.; Ramesh, A.; Ziegler, D.; Wu, J.; Winter, C.; Hesse, C.; Chen, M.; Sigler, E.; Litwin, M.; Gray, S.; Chess, B.; Clark, J.; Berner, C.; McCandlish, S.; Radford, A.; Sutskever, I.; and Amodei, D. 2020.
\newblock Language Models are Few-Shot Learners.
\newblock In Larochelle, H.; Ranzato, M.; Hadsell, R.; Balcan, M.; and Lin, H., eds., \emph{Advances in Neural Information Processing Systems}, volume~33, 1877--1901. Curran Associates, Inc.

\bibitem[{Chen et~al.(2023)Chen, Peng, srinivasan parthasarathy, and Ning}]{chen2023shapemol}
Chen, Z.; Peng, B.; srinivasan parthasarathy; and Ning, X. 2023.
\newblock Shape-conditioned 3{D} Molecule Generation via Equivariant Diffusion Models.
\newblock In \emph{NeurIPS 2023 Generative AI and Biology (GenBio) Workshop}.

\bibitem[{Chilingaryan et~al.(2024)Chilingaryan, Tamoyan, Tevosyan, Babayan, Hambardzumyan, Navoyan, Aghajanyan, Khachatrian, and Khondkaryan}]{chilingaryan2022bartsmiles}
Chilingaryan, G.; Tamoyan, H.; Tevosyan, A.; Babayan, N.; Hambardzumyan, K.; Navoyan, Z.; Aghajanyan, A.; Khachatrian, H.; and Khondkaryan, L. 2024.
\newblock {B}art{S}miles: Generative Masked Language Models for Molecular Representations.
\newblock \emph{Journal of Chemical Information and Modeling}, 64(15): 5832--5843.
\newblock PMID: 39054761.

\bibitem[{Chithrananda, Grand, and Ramsundar(2020)}]{chithrananda2020chemberta}
Chithrananda, S.; Grand, G.; and Ramsundar, B. 2020.
\newblock Chem{BERT}a: large-scale self-supervised pretraining for molecular property prediction.
\newblock \emph{arXiv preprint arXiv:2010.09885}.

\bibitem[{Christofidellis et~al.(2023)Christofidellis, Giannone, Born, Winther, Laino, and Manica}]{christofidellis2023unifying}
Christofidellis, D.; Giannone, G.; Born, J.; Winther, O.; Laino, T.; and Manica, M. 2023.
\newblock Unifying Molecular and Textual Representations via Multi-task Language Modelling.
\newblock In Krause, A.; Brunskill, E.; Cho, K.; Engelhardt, B.; Sabato, S.; and Scarlett, J., eds., \emph{Proceedings of the 40th International Conference on Machine Learning}, volume 202 of \emph{Proceedings of Machine Learning Research}, 6140--6157. PMLR.

\bibitem[{Chung et~al.(2024)Chung, Hou, Longpre, Zoph, Tay, Fedus, Li, Wang, Dehghani, Brahma, Webson, Gu, Dai, Suzgun, Chen, Chowdhery, Castro-Ros, Pellat, Robinson, Valter, Narang, Mishra, Yu, Zhao, Huang, Dai, Yu, Petrov, Chi, Dean, Devlin, Roberts, Zhou, Le, and Wei}]{flant5}
Chung, H.~W.; Hou, L.; Longpre, S.; Zoph, B.; Tay, Y.; Fedus, W.; Li, Y.; Wang, X.; Dehghani, M.; Brahma, S.; Webson, A.; Gu, S.~S.; Dai, Z.; Suzgun, M.; Chen, X.; Chowdhery, A.; Castro-Ros, A.; Pellat, M.; Robinson, K.; Valter, D.; Narang, S.; Mishra, G.; Yu, A.; Zhao, V.; Huang, Y.; Dai, A.; Yu, H.; Petrov, S.; Chi, E.~H.; Dean, J.; Devlin, J.; Roberts, A.; Zhou, D.; Le, Q.~V.; and Wei, J. 2024.
\newblock Scaling Instruction-Finetuned Language Models.
\newblock \emph{Journal of Machine Learning Research}, 25(70): 1--53.

\bibitem[{Degen et~al.(2008)Degen, Wegscheid-Gerlach, Zaliani, and Rarey}]{degen2008brics}
Degen, J.; Wegscheid-Gerlach, C.; Zaliani, A.; and Rarey, M. 2008.
\newblock On the Art of Compiling and Using 'Drug-Like' Chemical Fragment Spaces.
\newblock \emph{ChemMedChem}, 3(10): 1503--1507.

\bibitem[{Devlin et~al.(2019)Devlin, Chang, Lee, and Toutanova}]{devlin2019bert}
Devlin, J.; Chang, M.-W.; Lee, K.; and Toutanova, K. 2019.
\newblock {BERT}: Pre-training of Deep Bidirectional Transformers for Language Understanding.
\newblock In \emph{Proceedings of the 2019 Conference of the North {A}merican Chapter of the Association for Computational Linguistics: Human Language Technologies, Volume 1 (Long and Short Papers)}, 4171--4186. Minneapolis, Minnesota: Association for Computational Linguistics.

\bibitem[{Edwards et~al.(2022)Edwards, Lai, Ros, Honke, Cho, and Ji}]{edwards2022molt5}
Edwards, C.; Lai, T.; Ros, K.; Honke, G.; Cho, K.; and Ji, H. 2022.
\newblock Translation between Molecules and Natural Language.
\newblock In Goldberg, Y.; Kozareva, Z.; and Zhang, Y., eds., \emph{Proceedings of the 2022 Conference on Empirical Methods in Natural Language Processing}, 375--413. Abu Dhabi, United Arab Emirates: Association for Computational Linguistics.

\bibitem[{Ertl and Schuffenhauer(2009)}]{ertl2009sascore}
Ertl, P.; and Schuffenhauer, A. 2009.
\newblock Estimation of synthetic accessibility score of drug-like molecules based on molecular complexity and fragment contributions.
\newblock \emph{Journal of Cheminformatics}, 1(1): 8.

\bibitem[{Feng et~al.(2020)Feng, Guo, Tang, Duan, Feng, Gong, Shou, Qin, Liu, Jiang, and Zhou}]{feng2020codebert}
Feng, Z.; Guo, D.; Tang, D.; Duan, N.; Feng, X.; Gong, M.; Shou, L.; Qin, B.; Liu, T.; Jiang, D.; and Zhou, M. 2020.
\newblock {C}ode{BERT}: A Pre-Trained Model for Programming and Natural Languages.
\newblock In Cohn, T.; He, Y.; and Liu, Y., eds., \emph{Findings of the Association for Computational Linguistics: EMNLP 2020}, 1536--1547. Online: Association for Computational Linguistics.

\bibitem[{Fialková et~al.(2022)Fialková, Zhao, Papadopoulos, Engkvist, Bjerrum, Kogej, and Patronov}]{fialkova2022libinvent}
Fialková, V.; Zhao, J.; Papadopoulos, K.; Engkvist, O.; Bjerrum, E.~J.; Kogej, T.; and Patronov, A. 2022.
\newblock {L}ib{INVENT}: Reaction-based Generative Scaffold Decoration for in Silico Library Design.
\newblock \emph{Journal of Chemical Information and Modeling}, 62(9): 2046--2063.
\newblock PMID: 34460269.

\bibitem[{Flam-Shepherd and Aspuru-Guzik(2023)}]{flamshepherd2023xyz}
Flam-Shepherd, D.; and Aspuru-Guzik, A. 2023.
\newblock Language models can generate molecules, materials, and protein binding sites directly in three dimensions as {XYZ}, {CIF}, and {PDB} files.
\newblock arXiv:2305.05708.

\bibitem[{Flam-Shepherd, Zhu, and Aspuru-Guzik(2022)}]{flamshepherd2022lmmoldist}
Flam-Shepherd, D.; Zhu, K.; and Aspuru-Guzik, A. 2022.
\newblock Language models can learn complex molecular distributions.
\newblock \emph{Nature Communications}, 13(1): 3293.

\bibitem[{Francoeur et~al.(2020)Francoeur, Masuda, Sunseri, Jia, Iovanisci, Snyder, and Koes}]{francoeur2020crossdocked}
Francoeur, P.~G.; Masuda, T.; Sunseri, J.; Jia, A.; Iovanisci, R.~B.; Snyder, I.; and Koes, D.~R. 2020.
\newblock Three-Dimensional Convolutional Neural Networks and a Cross-Docked Data Set for Structure-Based Drug Design.
\newblock \emph{Journal of Chemical Information and Modeling}, 60(9): 4200--4215.
\newblock PMID: 32865404.

\bibitem[{Ganea et~al.(2021)Ganea, Pattanaik, Coley, Barzilay, Jensen, Green, and Jaakkola}]{ganea2021geomol}
Ganea, O.; Pattanaik, L.; Coley, C.; Barzilay, R.; Jensen, K.; Green, W.; and Jaakkola, T. 2021.
\newblock {G}eo{M}ol: Torsional Geometric Generation of Molecular 3D Conformer Ensembles.
\newblock In Ranzato, M.; Beygelzimer, A.; Dauphin, Y.; Liang, P.; and Vaughan, J.~W., eds., \emph{Advances in Neural Information Processing Systems}, volume~34, 13757--13769. Curran Associates, Inc.

\bibitem[{Geng and Liu(2023)}]{openlm2023openllama}
Geng, X.; and Liu, H. 2023.
\newblock Open{LL}a{M}A: An Open Reproduction of {LL}a{M}A.

\bibitem[{Guan et~al.(2023{\natexlab{a}})Guan, Peng, Jiang, Luo, Peng, and Ma}]{guan2023linkernet}
Guan, J.; Peng, X.; Jiang, P.; Luo, Y.; Peng, J.; and Ma, J. 2023{\natexlab{a}}.
\newblock {L}inker{N}et: Fragment Poses and Linker Co-Design with 3{D} Equivariant Diffusion.
\newblock In Oh, A.; Neumann, T.; Globerson, A.; Saenko, K.; Hardt, M.; and Levine, S., eds., \emph{Advances in Neural Information Processing Systems}, volume~36, 77503--77519. Curran Associates, Inc.

\bibitem[{Guan et~al.(2023{\natexlab{b}})Guan, Qian, Peng, Su, Peng, and Ma}]{guan2023targetdiff}
Guan, J.; Qian, W.~W.; Peng, X.; Su, Y.; Peng, J.; and Ma, J. 2023{\natexlab{b}}.
\newblock 3{D} Equivariant Diffusion for Target-Aware Molecule Generation and Affinity Prediction.
\newblock In \emph{The Eleventh International Conference on Learning Representations}.

\bibitem[{Guo et~al.(2021)Guo, Cai, Liu, Mu, Martin, and Hu}]{guo2021pct}
Guo, M.-H.; Cai, J.-X.; Liu, Z.-N.; Mu, T.-J.; Martin, R.~R.; and Hu, S.-M. 2021.
\newblock {PCT}: {P}oint cloud transformer.
\newblock \emph{Computational Visual Media}, 7(2): 187--199.

\bibitem[{Ho, Jain, and Abbeel(2020)}]{ho2020diffusion}
Ho, J.; Jain, A.; and Abbeel, P. 2020.
\newblock Denoising Diffusion Probabilistic Models.
\newblock In Larochelle, H.; Ranzato, M.; Hadsell, R.; Balcan, M.; and Lin, H., eds., \emph{Advances in Neural Information Processing Systems}, volume~33, 6840--6851. Curran Associates, Inc.

\bibitem[{Hoogeboom et~al.(2022)Hoogeboom, Satorras, Vignac, and Welling}]{hoogeboom2022edm}
Hoogeboom, E.; Satorras, V.~G.; Vignac, C.; and Welling, M. 2022.
\newblock Equivariant Diffusion for Molecule Generation in 3{D}.
\newblock In Chaudhuri, K.; Jegelka, S.; Song, L.; Szepesvari, C.; Niu, G.; and Sabato, S., eds., \emph{Proceedings of the 39th International Conference on Machine Learning}, volume 162 of \emph{Proceedings of Machine Learning Research}, 8867--8887. PMLR.

\bibitem[{Huang et~al.(2023)Huang, Zhang, Xu, and Wong}]{huang2023mdm}
Huang, L.; Zhang, H.; Xu, T.; and Wong, K.-C. 2023.
\newblock {MDM}: Molecular Diffusion Model for 3{D} Molecule Generation.
\newblock \emph{Proceedings of the AAAI Conference on Artificial Intelligence}, 37(4): 5105--5112.

\bibitem[{Huang et~al.(2022)Huang, Peng, Ma, and Zhang}]{huang223dlinker}
Huang, Y.; Peng, X.; Ma, J.; and Zhang, M. 2022.
\newblock 3{DL}inker: An E(3) Equivariant Variational Autoencoder for Molecular Linker Design.
\newblock In Chaudhuri, K.; Jegelka, S.; Song, L.; Szepesvari, C.; Niu, G.; and Sabato, S., eds., \emph{Proceedings of the 39th International Conference on Machine Learning}, volume 162 of \emph{Proceedings of Machine Learning Research}, 9280--9294. PMLR.

\bibitem[{Igashov et~al.(2024)Igashov, St{\"a}rk, Vignac, Schneuing, Satorras, Frossard, Welling, Bronstein, and Correia}]{igashov2022difflinker}
Igashov, I.; St{\"a}rk, H.; Vignac, C.; Schneuing, A.; Satorras, V.~G.; Frossard, P.; Welling, M.; Bronstein, M.; and Correia, B. 2024.
\newblock Equivariant 3D-conditional diffusion model for molecular linker design.
\newblock \emph{Nature Machine Intelligence}.

\bibitem[{Imrie et~al.(2020)Imrie, Bradley, van~der Schaar, and Deane}]{imrie2020delinker}
Imrie, F.; Bradley, A.~R.; van~der Schaar, M.; and Deane, C.~M. 2020.
\newblock Deep Generative Models for 3{D} Linker Design.
\newblock \emph{Journal of Chemical Information and Modeling}, 60(4): 1983--1995.
\newblock PMID: 32195587.

\bibitem[{Irwin et~al.(2020)Irwin, Tang, Young, Dandarchuluun, Wong, Khurelbaatar, Moroz, Mayfield, and Sayle}]{zinc20-dataset}
Irwin, J.~J.; Tang, K.~G.; Young, J.; Dandarchuluun, C.; Wong, B.~R.; Khurelbaatar, M.; Moroz, Y.~S.; Mayfield, J.~W.; and Sayle, R.~A. 2020.
\newblock {ZINC20} - {A} Free Ultralarge-Scale Chemical Database for Ligand Discovery.
\newblock \emph{J. Chem. Inf. Model.}, 60(12): 6065--6073.

\bibitem[{Irwin et~al.(2022)Irwin, Dimitriadis, He, and Bjerrum}]{irwin2022chemformer}
Irwin, R.; Dimitriadis, S.; He, J.; and Bjerrum, E.~J. 2022.
\newblock {C}hemformer: a pre-trained transformer for computational chemistry.
\newblock \emph{Machine Learning: Science and Technology}, 3(1): 015022.

\bibitem[{Jing et~al.(2022)Jing, Corso, Chang, Barzilay, and Jaakkola}]{jing2022torsionaldiffusion}
Jing, B.; Corso, G.; Chang, J.; Barzilay, R.; and Jaakkola, T. 2022.
\newblock Torsional Diffusion for Molecular Conformer Generation.
\newblock In Koyejo, S.; Mohamed, S.; Agarwal, A.; Belgrave, D.; Cho, K.; and Oh, A., eds., \emph{Advances in Neural Information Processing Systems}, volume~35, 24240--24253. Curran Associates, Inc.

\bibitem[{Koh, Salakhutdinov, and Fried(2023)}]{koh2023fromage}
Koh, J.~Y.; Salakhutdinov, R.; and Fried, D. 2023.
\newblock Grounding Language Models to Images for Multimodal Inputs and Outputs.
\newblock In Krause, A.; Brunskill, E.; Cho, K.; Engelhardt, B.; Sabato, S.; and Scarlett, J., eds., \emph{Proceedings of the 40th International Conference on Machine Learning}, volume 202 of \emph{Proceedings of Machine Learning Research}, 17283--17300. PMLR.

\bibitem[{Krenn et~al.(2020)Krenn, Häse, Nigam, Friederich, and Aspuru-Guzik}]{krenn2020selfies}
Krenn, M.; Häse, F.; Nigam, A.; Friederich, P.; and Aspuru-Guzik, A. 2020.
\newblock Self-referencing embedded strings ({SELFIES}): A 100\% robust molecular string representation.
\newblock \emph{Machine Learning: Science and Technology}, 1(4): 045024.

\bibitem[{Lacoste et~al.(2019)Lacoste, Luccioni, Schmidt, and Dandres}]{lacoste2019quantifying}
Lacoste, A.; Luccioni, A.; Schmidt, V.; and Dandres, T. 2019.
\newblock Quantifying the carbon emissions of machine learning.
\newblock \emph{arXiv preprint arXiv:1910.09700}.

\bibitem[{Landrum et~al.(2023)Landrum, Tosco, Kelley, Ric, Cosgrove, sriniker, gedeck, Vianello, NadineSchneider, Kawashima, Jones, N, Dalke, Cole, Swain, Turk, AlexanderSavelyev, Vaucher, Wójcikowski, Take, Scalfani, Probst, Ujihara, guillaume godin, Walker, Lehtivarjo, Pahl, Berenger, jasondbiggs, and strets123}]{landrum2023rdkit}
Landrum, G.; Tosco, P.; Kelley, B.; Ric; Cosgrove, D.; sriniker; gedeck; Vianello, R.; NadineSchneider; Kawashima, E.; Jones, G.; N, D.; Dalke, A.; Cole, B.; Swain, M.; Turk, S.; AlexanderSavelyev; Vaucher, A.; Wójcikowski, M.; Take, I.; Scalfani, V.~F.; Probst, D.; Ujihara, K.; guillaume godin; Walker, R.; Lehtivarjo, J.; Pahl, A.; Berenger, F.; jasondbiggs; and strets123. 2023.
\newblock rdkit/rdkit: 2023\_09\_3 (Q3 2023) Release.

\bibitem[{Landrum, Penzotti, and Putta(2006)}]{landrum2006feature}
Landrum, G.~A.; Penzotti, J.~E.; and Putta, S. 2006.
\newblock Feature-map vectors: a new class of informative descriptors for computational drug discovery.
\newblock \emph{Journal of computer-aided molecular design}, 20: 751--762.

\bibitem[{Lewis et~al.(2020)Lewis, Liu, Goyal, Ghazvininejad, Mohamed, Levy, Stoyanov, and Zettlemoyer}]{lewis2020bart}
Lewis, M.; Liu, Y.; Goyal, N.; Ghazvininejad, M.; Mohamed, A.; Levy, O.; Stoyanov, V.; and Zettlemoyer, L. 2020.
\newblock {BART}: Denoising Sequence-to-Sequence Pre-training for Natural Language Generation, Translation, and Comprehension.
\newblock In Jurafsky, D.; Chai, J.; Schluter, N.; and Tetreault, J., eds., \emph{Proceedings of the 58th Annual Meeting of the Association for Computational Linguistics}, 7871--7880. Online: Association for Computational Linguistics.

\bibitem[{Liang et~al.(2023)Liang, Zhang, Zhang, and Xie}]{liang2023drugchat}
Liang, Y.; Zhang, R.; Zhang, l.; and Xie, P. 2023.
\newblock DrugChat: Towards Enabling ChatGPT-Like Capabilities on Drug Molecule Graphs.
\newblock \emph{TechRxiv}.

\bibitem[{Lin et~al.(2023)Lin, Huang, Zhang, Liu, Wu, Li, Chen, and Li}]{lin2023d3fg}
Lin, H.; Huang, Y.; Zhang, O.; Liu, Y.; Wu, L.; Li, S.; Chen, Z.; and Li, S.~Z. 2023.
\newblock Functional-Group-Based Diffusion for Pocket-Specific Molecule Generation and Elaboration.
\newblock In Oh, A.; Neumann, T.; Globerson, A.; Saenko, K.; Hardt, M.; and Levine, S., eds., \emph{Advances in Neural Information Processing Systems}, volume~36, 34603--34626. Curran Associates, Inc.

\bibitem[{Liu, Chen, and Ding(2022)}]{liu2022tr}
Liu, L.; Chen, E.; and Ding, Y. 2022.
\newblock {TR}-{N}et: a transformer-based neural network for point cloud processing.
\newblock \emph{Machines}, 10(7): 517.

\bibitem[{Liu et~al.(2023)Liu, Tian, Lv, Li, and Wang}]{liu2023point}
Liu, Y.; Tian, B.; Lv, Y.; Li, L.; and Wang, F.-Y. 2023.
\newblock Point cloud classification using content-based transformer via clustering in feature space.
\newblock \emph{IEEE/CAA Journal of Automatica Sinica}.

\bibitem[{Livne et~al.(2024)Livne, Miftahutdinov, Tutubalina, Kuznetsov, Polykovskiy, Brundyn, Jhunjhunwala, Costa, Aliper, Aspuru-Guzik, and Zhavoronkov}]{nach0}
Livne, M.; Miftahutdinov, Z.; Tutubalina, E.; Kuznetsov, M.; Polykovskiy, D.; Brundyn, A.; Jhunjhunwala, A.; Costa, A.; Aliper, A.; Aspuru-Guzik, A.; and Zhavoronkov, A. 2024.
\newblock nach0: multimodal natural and chemical languages foundation model.
\newblock \emph{Chem. Sci.}, 15: 8380--8389.

\bibitem[{Lu et~al.(2019)Lu, Batra, Parikh, and Lee}]{lu2019vilbert}
Lu, J.; Batra, D.; Parikh, D.; and Lee, S. 2019.
\newblock {V}i{LBERT}: Pretraining Task-Agnostic Visiolinguistic Representations for Vision-and-Language Tasks.
\newblock In Wallach, H.; Larochelle, H.; Beygelzimer, A.; d\textquotesingle Alch\'{e}-Buc, F.; Fox, E.; and Garnett, R., eds., \emph{Advances in Neural Information Processing Systems}, volume~32. Curran Associates, Inc.

\bibitem[{Lu and Zhang(2022)}]{lu2022t5chem}
Lu, J.; and Zhang, Y. 2022.
\newblock Unified Deep Learning Model for Multitask Reaction Predictions with Explanation.
\newblock \emph{Journal of Chemical Information and Modeling}, 62(6): 1376--1387.
\newblock PMID: 35266390.

\bibitem[{Luo et~al.(2022)Luo, Chen, Xu, Zheng, Liu, Wang, and He}]{luo2022one}
Luo, S.; Chen, T.; Xu, Y.; Zheng, S.; Liu, T.-Y.; Wang, L.; and He, D. 2022.
\newblock One transformer can understand both 2d \& 3d molecular data.
\newblock In \emph{The Eleventh International Conference on Learning Representations}.

\bibitem[{Luo et~al.(2021)Luo, Guan, Ma, and Peng}]{luo2021ar}
Luo, S.; Guan, J.; Ma, J.; and Peng, J. 2021.
\newblock A 3{D} Generative Model for Structure-Based Drug Design.
\newblock In Ranzato, M.; Beygelzimer, A.; Dauphin, Y.; Liang, P.; and Vaughan, J.~W., eds., \emph{Advances in Neural Information Processing Systems}, volume~34, 6229--6239. Curran Associates, Inc.

\bibitem[{O'Boyle et~al.(2011)O'Boyle, Banck, James, Morley, Vandermeersch, and Hutchison}]{oboyle2011openbabel}
O'Boyle, N.~M.; Banck, M.; James, C.~A.; Morley, C.; Vandermeersch, T.; and Hutchison, G.~R. 2011.
\newblock {O}pen {B}abel: An open chemical toolbox.
\newblock \emph{Journal of Cheminformatics}, 3(1): 33.

\bibitem[{Pei et~al.(2023)Pei, Zhang, Zhu, Wu, Gao, Wu, Xia, and Yan}]{pei2023biot5}
Pei, Q.; Zhang, W.; Zhu, J.; Wu, K.; Gao, K.; Wu, L.; Xia, Y.; and Yan, R. 2023.
\newblock {B}io{T}5: Enriching Cross-modal Integration in Biology with Chemical Knowledge and Natural Language Associations.
\newblock In Bouamor, H.; Pino, J.; and Bali, K., eds., \emph{Proceedings of the 2023 Conference on Empirical Methods in Natural Language Processing}, 1102--1123. Singapore: Association for Computational Linguistics.

\bibitem[{Peng et~al.(2023)Peng, Guan, Liu, and Ma}]{peng2023moldiff}
Peng, X.; Guan, J.; Liu, Q.; and Ma, J. 2023.
\newblock {M}ol{D}iff: Addressing the Atom-Bond Inconsistency Problem in 3{D} Molecule Diffusion Generation.
\newblock In Krause, A.; Brunskill, E.; Cho, K.; Engelhardt, B.; Sabato, S.; and Scarlett, J., eds., \emph{Proceedings of the 40th International Conference on Machine Learning}, volume 202 of \emph{Proceedings of Machine Learning Research}, 27611--27629. PMLR.

\bibitem[{Peng et~al.(2022)Peng, Luo, Guan, Xie, Peng, and Ma}]{peng2022pockettomol}
Peng, X.; Luo, S.; Guan, J.; Xie, Q.; Peng, J.; and Ma, J. 2022.
\newblock {P}ocket2{M}ol: Efficient Molecular Sampling Based on 3{D} Protein Pockets.
\newblock In Chaudhuri, K.; Jegelka, S.; Song, L.; Szepesvari, C.; Niu, G.; and Sabato, S., eds., \emph{Proceedings of the 39th International Conference on Machine Learning}, volume 162 of \emph{Proceedings of Machine Learning Research}, 17644--17655. PMLR.

\bibitem[{Pinheiro et~al.(2023)Pinheiro, Rackers, joseph Kleinhenz, Maser, Mahmood, Watkins, Ra, Sresht, and Saremi}]{pinheiro2023voxmol}
Pinheiro, P.~O.; Rackers, J.; joseph Kleinhenz; Maser, M.; Mahmood, O.; Watkins, A.~M.; Ra, S.; Sresht, V.; and Saremi, S. 2023.
\newblock 3{D} molecule generation by denoising voxel grids.
\newblock In \emph{Thirty-seventh Conference on Neural Information Processing Systems}.

\bibitem[{Polykovskiy et~al.(2020)Polykovskiy, Zhebrak, Sanchez-Lengeling, Golovanov, Tatanov, Belyaev, Kurbanov, Artamonov, Aladinskiy, Veselov, Kadurin, Johansson, Chen, Nikolenko, Aspuru-Guzik, and Zhavoronkov}]{polykovskiy2020moses}
Polykovskiy, D.; Zhebrak, A.; Sanchez-Lengeling, B.; Golovanov, S.; Tatanov, O.; Belyaev, S.; Kurbanov, R.; Artamonov, A.; Aladinskiy, V.; Veselov, M.; Kadurin, A.; Johansson, S.; Chen, H.; Nikolenko, S.; Aspuru-Guzik, A.; and Zhavoronkov, A. 2020.
\newblock {M}olecular {S}ets ({MOSES}): A Benchmarking Platform for Molecular Generation Models.
\newblock \emph{Frontiers in Pharmacology}, 11.

\bibitem[{Pracht, Bohle, and Grimme(2020)}]{pracht2020automated}
Pracht, P.; Bohle, F.; and Grimme, S. 2020.
\newblock Automated exploration of the low-energy chemical space with fast quantum chemical methods.
\newblock \emph{Physical Chemistry Chemical Physics}, 22(14): 7169--7192.

\bibitem[{Putta, Landrum, and Penzotti(2005)}]{putta2005conformation}
Putta, S.; Landrum, G.~A.; and Penzotti, J.~E. 2005.
\newblock Conformation mining: an algorithm for finding biologically relevant conformations.
\newblock \emph{Journal of medicinal chemistry}, 48(9): 3313--3318.

\bibitem[{Radford et~al.(2023)Radford, Kim, Xu, Brockman, Mcleavey, and Sutskever}]{radford2023whisper}
Radford, A.; Kim, J.~W.; Xu, T.; Brockman, G.; Mcleavey, C.; and Sutskever, I. 2023.
\newblock Robust Speech Recognition via Large-Scale Weak Supervision.
\newblock In Krause, A.; Brunskill, E.; Cho, K.; Engelhardt, B.; Sabato, S.; and Scarlett, J., eds., \emph{Proceedings of the 40th International Conference on Machine Learning}, volume 202 of \emph{Proceedings of Machine Learning Research}, 28492--28518. PMLR.

\bibitem[{Raffel et~al.(2020)Raffel, Shazeer, Roberts, Lee, Narang, Matena, Zhou, Li, and Liu}]{raffel2020t5}
Raffel, C.; Shazeer, N.; Roberts, A.; Lee, K.; Narang, S.; Matena, M.; Zhou, Y.; Li, W.; and Liu, P.~J. 2020.
\newblock Exploring the Limits of Transfer Learning with a Unified Text-to-Text Transformer.
\newblock \emph{Journal of Machine Learning Research}, 21(140): 1--67.

\bibitem[{Ragoza, Masuda, and Koes(2022)}]{ragoza2022ligan}
Ragoza, M.; Masuda, T.; and Koes, D.~R. 2022.
\newblock Generating 3{D} molecules conditional on receptor binding sites with deep generative models.
\newblock \emph{Chem. Sci.}, 13: 2701--2713.

\bibitem[{Ren et~al.(2019)Ren, Ruan, Tan, Qin, Zhao, Zhao, and Liu}]{ren2019fastspeech}
Ren, Y.; Ruan, Y.; Tan, X.; Qin, T.; Zhao, S.; Zhao, Z.; and Liu, T.-Y. 2019.
\newblock {F}ast{S}peech: Fast, Robust and Controllable Text to Speech.
\newblock In Wallach, H.; Larochelle, H.; Beygelzimer, A.; d\textquotesingle Alch\'{e}-Buc, F.; Fox, E.; and Garnett, R., eds., \emph{Advances in Neural Information Processing Systems}, volume~32. Curran Associates, Inc.

\bibitem[{Ross et~al.(2022)Ross, Belgodere, Chenthamarakshan, Padhi, Mroueh, and Das}]{ross2022molformer}
Ross, J.; Belgodere, B.; Chenthamarakshan, V.; Padhi, I.; Mroueh, Y.; and Das, P. 2022.
\newblock Large-scale chemical language representations capture molecular structure and properties.
\newblock \emph{Nature Machine Intelligence}, 4(12): 1256--1264.

\bibitem[{Sch\"{u}tt et~al.(2017)Sch\"{u}tt, Kindermans, Sauceda~Felix, Chmiela, Tkatchenko, and M\"{u}ller}]{schutt2017schnet}
Sch\"{u}tt, K.; Kindermans, P.-J.; Sauceda~Felix, H.~E.; Chmiela, S.; Tkatchenko, A.; and M\"{u}ller, K.-R. 2017.
\newblock {S}ch{N}et: A continuous-filter convolutional neural network for modeling quantum interactions.
\newblock In Guyon, I.; Luxburg, U.~V.; Bengio, S.; Wallach, H.; Fergus, R.; Vishwanathan, S.; and Garnett, R., eds., \emph{Advances in Neural Information Processing Systems}, volume~30. Curran Associates, Inc.

\bibitem[{Sohl-Dickstein et~al.(2015)Sohl-Dickstein, Weiss, Maheswaranathan, and Ganguli}]{sohldickstein2015diffusion}
Sohl-Dickstein, J.; Weiss, E.; Maheswaranathan, N.; and Ganguli, S. 2015.
\newblock Deep Unsupervised Learning using Nonequilibrium Thermodynamics.
\newblock In Bach, F.; and Blei, D., eds., \emph{Proceedings of the 32nd International Conference on Machine Learning}, volume~37 of \emph{Proceedings of Machine Learning Research}, 2256--2265. Lille, France: PMLR.

\bibitem[{Sun et~al.(2019)Sun, Myers, Vondrick, Murphy, and Schmid}]{sun2019videobert}
Sun, C.; Myers, A.; Vondrick, C.; Murphy, K.; and Schmid, C. 2019.
\newblock {V}ideo{BERT}: A Joint Model for Video and Language Representation Learning.
\newblock In \emph{2019 IEEE/CVF International Conference on Computer Vision (ICCV)}, 7463--7472. Los Alamitos, CA, USA: IEEE Computer Society.

\bibitem[{Tang et~al.(2024)Tang, Tran, Tan, and Gerstein}]{tang2023mollm}
Tang, X.; Tran, A.; Tan, J.; and Gerstein, M. 2024.
\newblock MolLM: A Unified Language Model to Integrate Biomedical Text with 2D and 3D Molecular Representations.
\newblock \emph{Bioinformatics}.

\bibitem[{Trott and Olson(2010)}]{trott2010vina}
Trott, O.; and Olson, A.~J. 2010.
\newblock {A}uto{D}ock {V}ina: Improving the speed and accuracy of docking with a new scoring function, efficient optimization, and multithreading.
\newblock \emph{Journal of Computational Chemistry}, 31(2): 455--461.

\bibitem[{Vaswani et~al.(2017)Vaswani, Shazeer, Parmar, Uszkoreit, Jones, Gomez, Kaiser, and Polosukhin}]{vaswani2017transformer}
Vaswani, A.; Shazeer, N.; Parmar, N.; Uszkoreit, J.; Jones, L.; Gomez, A.~N.; Kaiser, L.~u.; and Polosukhin, I. 2017.
\newblock Attention is All you Need.
\newblock In Guyon, I.; Luxburg, U.~V.; Bengio, S.; Wallach, H.; Fergus, R.; Vishwanathan, S.; and Garnett, R., eds., \emph{Advances in Neural Information Processing Systems}, volume~30. Curran Associates, Inc.

\bibitem[{Weininger(1988)}]{weininger1988smiles}
Weininger, D. 1988.
\newblock {SMILES}, a chemical language and information system. 1. Introduction to methodology and encoding rules.
\newblock \emph{Journal of Chemical Information and Computer Sciences}, 28(1): 31--36.

\bibitem[{Xie et~al.(2024)Xie, Chen, Lei, and Yang}]{xie2024diffdec}
Xie, J.; Chen, S.; Lei, J.; and Yang, Y. 2024.
\newblock {D}iff{D}ec: Structure-Aware Scaffold Decoration with an End-to-End Diffusion Model.
\newblock \emph{Journal of Chemical Information and Modeling}, 64(7): 2554--2564.
\newblock PMID: 38267393.

\bibitem[{Xu et~al.(2022)Xu, Yu, Song, Shi, Ermon, and Tang}]{xu2022geodiff}
Xu, M.; Yu, L.; Song, Y.; Shi, C.; Ermon, S.; and Tang, J. 2022.
\newblock {G}eo{D}iff: A Geometric Diffusion Model for Molecular Conformation Generation.
\newblock In \emph{International Conference on Learning Representations}.

\bibitem[{Yu et~al.(2022)Yu, Tang, Rao, Huang, Zhou, and Lu}]{yu2022pointbert}
Yu, X.; Tang, L.; Rao, Y.; Huang, T.; Zhou, J.; and Lu, J. 2022.
\newblock {P}oint-{BERT}: Pre-training 3{D} Point Cloud Transformers with Masked Point Modeling.
\newblock In \emph{2022 IEEE/CVF Conference on Computer Vision and Pattern Recognition (CVPR)}, 19291--19300.

\bibitem[{Zhang et~al.(2023)Zhang, Min, Zheng, and Liu}]{zhang2023flag}
Zhang, Z.; Min, Y.; Zheng, S.; and Liu, Q. 2023.
\newblock Molecule Generation For Target Protein Binding with Structural Motifs.
\newblock In \emph{The Eleventh International Conference on Learning Representations}.

\bibitem[{Zhao et~al.(2021)Zhao, Jiang, Jia, Torr, and Koltun}]{zhao2021pointtransformer}
Zhao, H.; Jiang, L.; Jia, J.; Torr, P.; and Koltun, V. 2021.
\newblock {P}oint {T}ransformer.
\newblock In \emph{2021 IEEE/CVF International Conference on Computer Vision (ICCV)}, 16239--16248.

\end{thebibliography}

\newpage~\newpage
\appendix
\section{Limitations}\label{app:limitations}
First, \ourmodel{} inherits limitations from its LM component. In this work, we utilized nach0, which was pre-trained on SMILES. A notable drawback of the SMILES format is the lack of a one-to-one correspondence between molecules and SMILES strings. A molecule can have multiple SMILES representations due to variations in the starting atom, molecular graph traversal, and kekulization. Second, scaling \ourmodel{} for long chemical sequences poses challenges due to the quadratic nature of the attention mechanism. Finally, it is important to recognize that the academic datasets used in this study mainly include existing drugs and known chemical probes, covering only a small portion of the vast predicted chemical space. Moreover, they do not account for testing on novel chemical diversity that differs from molecules reported in the literature. 
\section{Dataset Statistics and Examples of Inputs and Outputs}\label{app:datasets}
We employ several datasets to evaluate and benchmark our generative models for different molecular generation tasks. These datasets provide comprehensive and high-quality molecular data, facilitating tasks such as property prediction, conformation generation, and drug design. Below are detailed descriptions of the datasets used. Table \ref{tab:input_output_examples} shows examples of inputs and outputs for each task. We utilize listed datasets in the pretraining stage. 

It is important to mention that we cope with imbalanced dataset sizes by utilizing a balanced batch construction procedure - we include the samples from different tasks with the same probability of forming the final training batch.

\subsection{GEOM}
For spatial molecular distribution learning and conformation generation tasks, we use the high-quality GEOM-Drugs dataset \cite{axelrod2022geom}, which provides gold-standard conformer ensembles generated with metadynamics in CREST \cite{pracht2020automated}. 
The Geometric Ensemble Of Molecules (GEOM) dataset is an extensive collection of molecular conformations generated using advanced sampling and semi-empirical density functional theory (DFT). The dataset contains 37 million molecular conformations for over 450,000 molecules. The GEOM dataset includes conformers for 133,000 species from QM9 and 317,000 species with experimental data related to biophysics, physiology, and physical chemistry. The dataset also contains ensembles of 1,511 species with BACE-1 inhibition data labeled with high-quality DFT free energies in an implicit water solvent, and 534 ensembles are further optimized with DFT. The dataset can be used for training and benchmarking models in tasks such as molecular property prediction, conformation generation, and drug design.

\subsection{ZINC dataset}
We employ a subset of the ZINC dataset for the linker design task, consisting of 250,000 randomly selected molecules with low-energy conformations generated using RDKit \cite{landrum2023rdkit}. The ZINC dataset is a comprehensive collection of commercially available chemical compounds, primarily curated for virtual screening applications. It contains molecular graphs, each with up to 38 heavy atoms. This dataset is extensively used in computational chemistry and drug discovery research for tasks such as molecular property prediction. It provides a diverse and well-characterized set of molecules, facilitating the development of models that produce chemically valid, unique, and novel compounds with desirable properties.

In the linker design task, each molecule undergoes fragmentation by breaking all double cuts of acyclic single bonds, and the resulting splits are filtered according to a set of predefined rules. As a result of this fragmentation process, one molecule can yield various combinations of two fragments with a linker in between. In our experiments, we focus our efforts on generating the linker without specifying anchor points (connecting points in the input fragment). We adopt precomputed linker/fragment split from \cite{igashov2022difflinker}.

\subsection{MOSES}
We utilize the MOSES dataset \cite{polykovskiy2020moses} for the shape-conditioned generation task. Initially, this dataset is filtered to retain only molecules containing the 100 most prevalent fragments, and then 3D conformers are generated for the remaining molecules using RDKit \cite{landrum2023rdkit}. The MOSES dataset is a benchmarking platform that offers an extensive dataset and a suite of metrics for evaluating generative models in the context of unconditional molecular generation tasks. The dataset includes nearly 2 million molecular samples, filtered using MCF, PAINS, and other rules to ensure quality. 

In the shape-conditioned generation, we use conformations computed with RDKit \cite{landrum2023rdkit} and provided in \cite{adams2023squid} work.

\subsection{CrossDocked}

For the scaffold-decoration and pocket-conditioned generation tasks, we employ CrossDocked \cite{francoeur2020crossdocked} dataset. It comprises 22.5 million protein-ligand complexes. It was preprocessed by leaving only 100,000 ligand-protein pairs with the highest binding affinity and clipping protein pocket region within 10{\AA} of ligand. Further, LibINVENT \cite{fialkova2022libinvent} slicing method with 37 customized reaction-based rules is applied on ligands to obtain scaffolds and R-groups.

\begin{table*}
\begin{center}
\begin{tabular}{|c|c|c|c|}
\hline
Task name & Input text & Input point cloud & Output text \\
\hline
\textbf{\shortstack{Distribution\\learning}} & \texttt{\shortstack{Generate \\ molecular \\ 3d structure\\from GEOM}} & \texttt{No input PC}  & \texttt{\shortstack{c1(c(ccc(Br)c1)O)C\#N|\\C -0.04 0.53 -0.51|C -1.24 0.22 0.15|\\C -1.23 -0.77 1.13|C -0.07 -1.44 1.45|\\C 1.11 -1.12 0.78|Br 2.71 -2.05 1.23|\\C 1.13 -0.15 -0.19|O -2.36 0.90 -0.19|\\C -0.02 1.53 -1.51|N 0.02 2.34 -2.33}} \\
\hline
\textbf{\shortstack{Conformation\\generation}} & \texttt{\shortstack{Generate \\ molecular \\ 3d structure\\for\\ S=c1ccccn1[O-]}} & \texttt{No input PC}  & \texttt{\shortstack{S -0.39 -2.50 0.33|C 0.03 -0.90 0.02|\\C 1.19 -0.56 -0.69|C 1.54 0.74 -0.95|\\C 0.71 1.76 -0.49|C -0.43 1.41 0.20|\\N -0.79 0.14 0.46|O -1.86 -0.09 1.11}} \\
\hline
\textbf{\shortstack{Shape-\\conditioned\\generation}} & \texttt{\shortstack{Generate \\ molecular \\ 3d structure\\for shape}} & \includegraphics[width=0.2\linewidth]{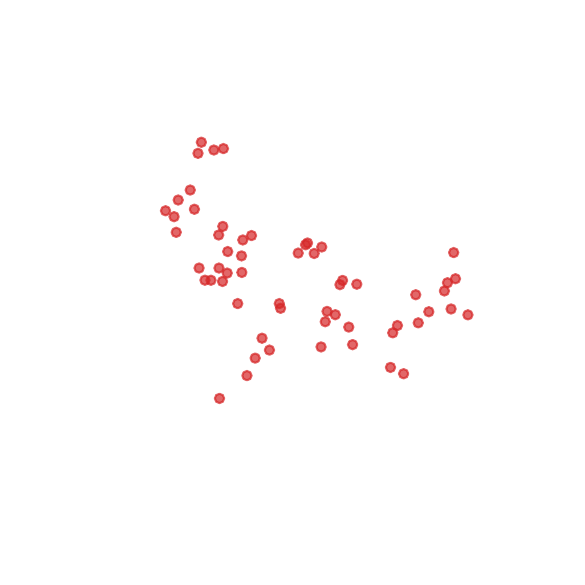} & \texttt{\shortstack{C(O)c1cc(c(c(Br)c1)OC)Br|\\C -2.29 1.45 -1.64|O -1.74 2.64 -2.19|\\C -1.24 0.71 -0.86|C -1.57 0.06 0.34|\\C -0.60 -0.66 1.05|C 0.71 -0.73 0.58|\\C 1.04 -0.13 -0.64|Br 2.77 -0.28 -1.41|\\C 0.07 0.59 -1.35|O 1.66 -1.46 1.27|\\C 2.35 -0.63 2.20|Br -1.15 -1.55 2.64}} \\
\hline
\textbf{\shortstack{Linker\\design}} & \texttt{\shortstack{Generate\\linker}} & \includegraphics[width=0.2\linewidth]{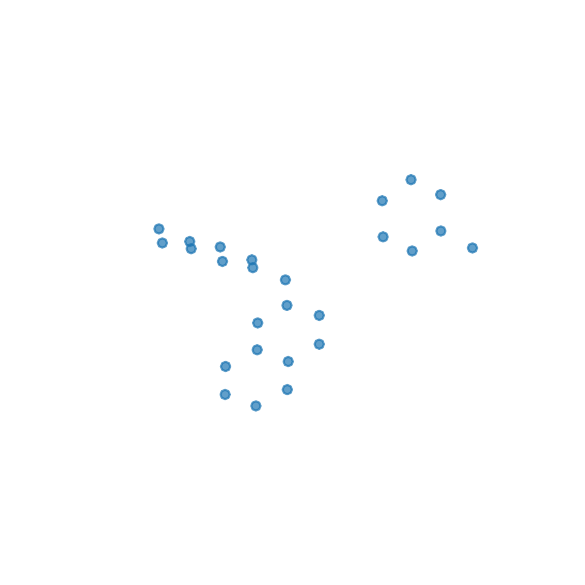} & \texttt{\shortstack{*NC(=O)*|\\ {*} 3.17 1.82 -0.34|N 2.05 1.00 -0.70|\\C 2.01 -0.43 -0.56|O 3.07 -1.08 -0.33|\\ {*} 0.75 -1.17 -0.76}} \\
\hline
\textbf{\shortstack{Scaffold\\decoration}} & \texttt{\shortstack{Generate\\decoration}} & \includegraphics[width=0.2\linewidth]{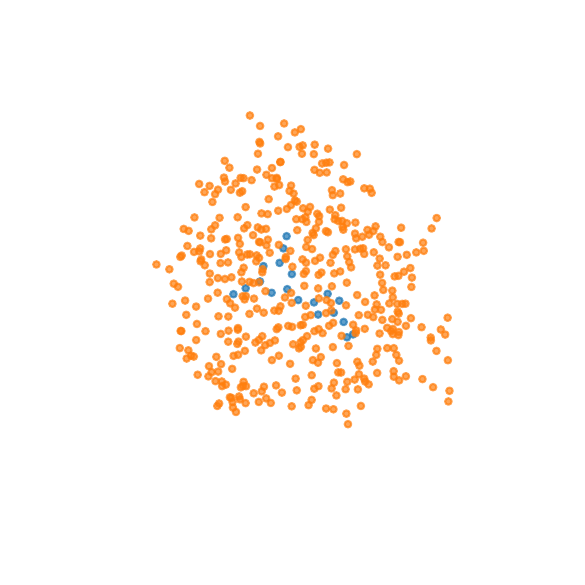} & \texttt{\shortstack{C(C)*.C(C)*|\\C 2.50 0.39 1.83|C 3.94 0.49 2.38|\\ {*} 2.38 -0.78 0.88|C 0.66 -3.60 -1.12|\\C 0.30 -4.67 -0.27|{*} 1.58 -2.55 -0.43}} \\
\hline
\textbf{\shortstack{Pocket-\\conditioned\\generation}} & \texttt{\shortstack{Generate\\molecular\\3d structure\\for pocket}} & \includegraphics[width=0.2\linewidth]{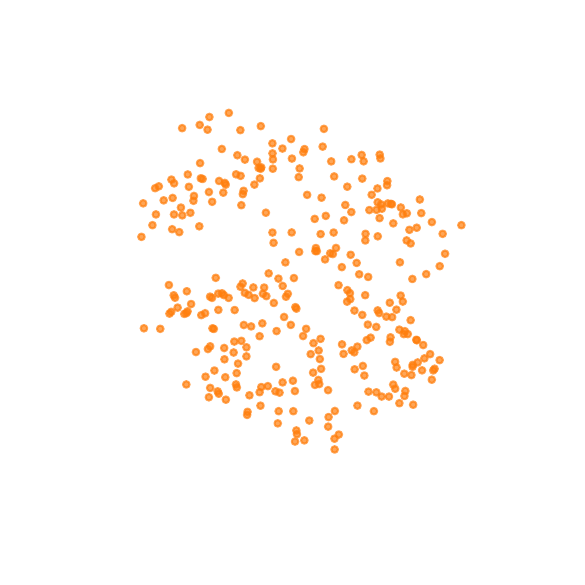} & \texttt{\shortstack{C1=NC(c2ccccc2)C=N1|\\C -2.54 1.46 -1.93|N -1.96 0.35 -1.33|\\C -0.73 0.73 -0.94|C 0.25 -0.18 -0.24|\\C 0.44 -1.48 -0.68|C 1.35 -2.37 -0.04|\\C 2.08 -1.87 1.08|C 1.91 -0.55 1.56|\\C 0.98 0.33 0.88|C -0.62 2.01 -1.30|\\N -1.74 2.39 -1.89}} \\
\hline
\end{tabular}
\end{center}
\caption{Examples of Inputs and Outputs. ``\texttt{No input PC}'' indicates the absence of an input point cloud.}
\label{tab:input_output_examples}
\end{table*}
\section{Additional Samples Visualizations}

\begin{figure*}[t]
\begin{center}
\centerline{\includegraphics[width=1.75\columnwidth]{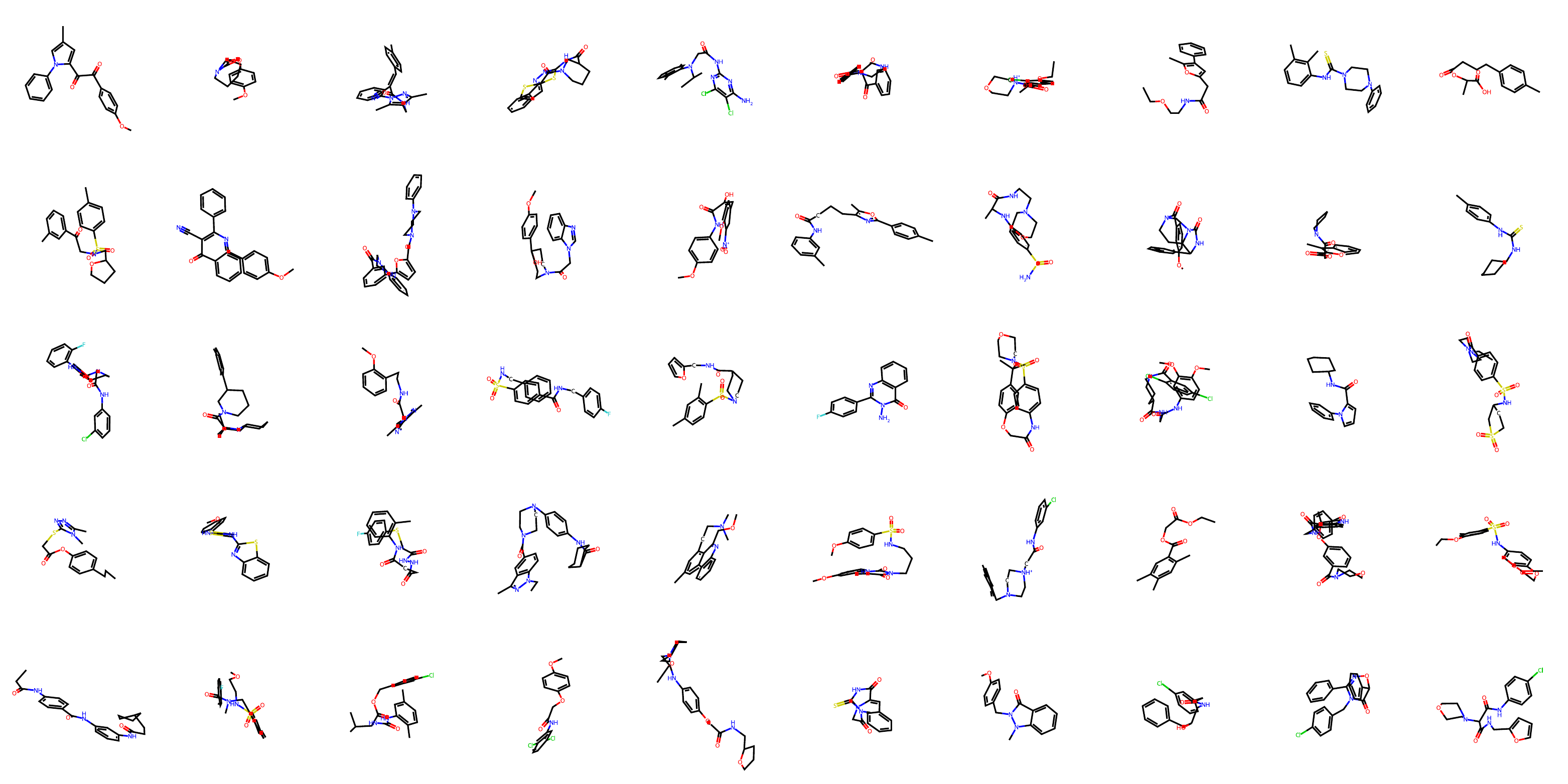}}
\caption{Additional samples for spatial molecular generation task on GEOM for Multi-task \ourmodel{}.}
\label{fig:samples_distribution_learning_3d}
\end{center}
\end{figure*}

Figure \ref{fig:samples_distribution_learning_3d} illustrates samples from the GEOM dataset, further demonstrating the model's versatility and capability in handling different types of molecular data. This visual representation aids in better understanding the molecular architecture and highlights the capability of our generative model to produce diverse and complex chemical entities.
\section{Evaluation Metrics}\label{app:metrics}

\subsection{Molecular distribution learning} 
Following \cite{peng2023moldiff}, we evaluate the valid and complete molecules from five perspectives: basic, drug-likeness, 3D structures, bonds, and rings.

\paragraph{Basic} We provide validity, uniqueness, novelty, and diversity metrics to evaluate the model's ability to generate physically and structurally plausible molecular spatial structures without mode collapsing and overfitting on the training set.

\paragraph{Drug-likeness} We assessed the drug-likeness of the generated molecules using the following metrics: (1) QED, which stands for quantitative estimation of drug-likeness; (2) SA, representing the synthetic accessibility score (higher values indicate easier synthesis); (3) Lipinski, which measures the number of Lipinski’s rule of five criteria the molecule meets.

\paragraph{3D structures} A key distinction between 3D generation and 2D molecule graph generation is the determination of atom positions, making it essential to measure their accuracy. We identified the most frequent bonds, bond pairs, and bond triplets in the validation set. We measured the bond lengths, bond angles, and dihedral angles for both the generated molecules and those in the validation set. We used the Jensen-Shannon (JS) divergence to quantify the differences in the distributions between the generated molecules and the validation set.

\paragraph{Bonds} We examined the bond-related properties of the generated molecules. Initially, we compared the distribution of bond counts per atom between the generated molecules and the validation set to determine if the models produced an excessive or insufficient number of bonds. Next, we analyzed the distributions of various bond types, including basic bond types (single, double, triple, and aromatic bonds) and the frequent bond types, bond pairs, and bond triplets used in the 3D structure evaluation.

\paragraph{Rings} We extended our bond analysis to include rings. First, we compared the distribution of ring counts per molecule between the generated molecules and the validation set. We compared the distributions of counts for rings of various sizes (n-sized rings) between the generated molecules and the validation set using JS divergence, averaging the JS divergence for $n \in \{3, 4, \ldots, 9\}$.

\subsection{Conformation generation} As evaluation metrics for conformer generation, we adopted the approach from \cite{jing2022torsionaldiffusion}, utilizing Average Minimum RMSD (AMR) and Coverage (COV) for Precision (P) and Recall (R) measured when generating twice as many conformers as provided by CREST. For \(K = 2L\), let \(\{C_l^*\}_{l \in [1, L]}\) and \(\{C_k\}_{k \in [1, K]}\) represent the sets of ground truth and generated conformers, respectively:

$$
\text{COV-R} := \frac{\left| \{l \in [1..L] : \exists k \in [1..K], \text{RMSD}(C_k, C_l^*) < \delta \} \right|}{L} 
$$

$$
\text{AMR-R} := \frac{\sum_{l \in [1..L]} \min_{k \in [1..K]} \text{RMSD}(C_k, C_l^*)}{L} 
$$

where $\delta$ is the coverage threshold. Following Tor.Diff \cite{jing2022torsionaldiffusion}, we utilize $\delta = 0.75$ {\AA} for our experiments.  Precision metrics are obtained by swapping the ground truth and generated conformers.

\subsection{Linker Generation}

We follow the evaluation procedure established in \cite{imrie2020delinker} and compute the Validity, Uniqueness, Recovery, RMSD, and  $SC$ of the samples. Subsequently, we ascertain whether the generated linkers adhere to the 2D filters utilized in generating the ZINC training set \cite{imrie2020delinker}.

\textbf{Validity} For validity calculation, we implement sanitization and additionally verify that the molecule includes all atoms from the fragments. For all other metrics, we consider only a subset of valid samples. 

\textbf{Uniqueness} We compare SMILES of whole molecules and count the number of unique molecules sampled for each input pair of fragments. 

\textbf{RMSD} To evaluate the 3D conformations of the sampled versus ground-truth molecules, we determine the Root Mean Squared Deviation (RMSD) between the coordinates of the generated linkers and those of the real molecules. To compute RMSD, we align the generated linker with the corresponding ground-truth molecules using RDKit function \texttt{rdkit.Chem.rdMolAlign}, which returns the optimal RMSD for aligning two molecules. 

$\textbf{SC}$ We utilize the $SC$ metric to measure the chemical similarity between the generated molecules and the ground-truth counterparts \cite{putta2005conformation, landrum2006feature}.

\textbf{2D filters} We applied 2D filters, including synthetic accessibility \cite{ertl2009sascore}, ring aromaticity (RA), and pan-assay interference compounds (PAINS) \cite{baell2010pains}, to create the ZINC and CASF datasets. The RA filter ensures correct covalent bond orders in ring structures, while the PAINS filter detects compounds prone to producing false-positive results in high-throughput screenings \cite{baell2010pains}.

\subsection{Scaffold Decoration}

We employ the same metrics for scaffold decoration as for linker generation. Additionally, we incorporate the Vina Scores and High Affinity metrics.

\textbf{QVina score} This score is calculated by QVina software \cite{alhossary2015qvina} to measure the binding affinity. QVina is designed to perform the same function as AutoDock Vina—predicting the binding affinity and orientation of a ligand to a protein receptor—but with optimizations that allow it to perform these calculations more quickly. The QVina scores serve as a metrics to evaluate the binding affinity between a ligand and a receptor in docking simulations. It quantitatively represents the predicted binding energy, with lower scores indicating stronger binding affinity. This score is essential in assessing the likelihood of successful ligand-receptor interactions and is particularly valuable in computational drug discovery and virtual screening efforts.

\subsection{Shape-conditioned Generation}

To evaluate shape-conditioned generation tasks, we utilize graph similarity ($\textbf{Sim}_{\textbf{G}}$) and shape similarity ($\textbf{Sim}_{\textbf{S}}$) metrics. 

$\textbf{Sim}_{\textbf{G}}$ We define graph (chemical) similarity, $sim_{G}  \in [0, 1] $, between two molecules as the Tanimoto similarity, computed by RDKit with default settings using 2048-bit fingerprints. 

$\textbf{Sim}_{\textbf{S}}$ We utilize shape similarity metric from ESP-Sim package \cite{bolcato2022shapesim}. 

\subsection{Pocket-conditioned generation}

\textbf{Vina Dock} For the pocket-conditioned generation tasks, we adopt the Vina Dock. To calculate the Vina Dock values, we utilize AutoDock Vina \cite{trott2010vina}, a widely employed software for molecular docking studies.

\textbf{High Affinity metric} The High Affinity metric is used to measure how strongly a ligand binds to a receptor in molecular docking or biochemical assays in comparison with reference ligand. It helps predict successful interactions between molecules. It considers factors like the stability of the ligand-receptor complex, the strength of interactions between them (like hydrogen bonds), and the overall energy of the interaction. We compared Vina Dock values for generated and reference molecules and provide percentage of cases when generated molecules binds better than reference molecules.
\section{Additional Related Work} \label{app:re}

\paragraph{Language Models in Chemistry}

The sequential nature of molecules enables the use of Transformer models and pre-training methods with models like ChemBERTa \cite{chithrananda2020chemberta}, T5Chem \cite{lu2022t5chem}, ChemFormer \cite{irwin2022chemformer}, and BARTSmiles \cite{chilingaryan2022bartsmiles}, which utilize masked language modeling for molecular SMILES. Incorporating both chemical and linguistic knowledge into LMs presents a novel and complex challenge. Recent advancements have introduced domain-specific LMs \cite{edwards2022molt5,christofidellis2023unifying,pei2023biot5}, based on T5, designed specifically to address this challenge. MolT5 \cite{edwards2022molt5} uses initial pre-training on a collection of molecule SMILES and texts, followed by single-task fine-tuning for molecule captioning (molecule→text) and text-based molecule generation (text→molecule) tasks. On the other hand, Text+Chem T5 \cite{christofidellis2023unifying} is a cross-domain, multi-task T5 model fine-tuned on five tasks, including forward reaction prediction and retro-synthesis. Another recent model, multi-domain Nach0 \cite{nach0}, undergoes fine-tuning on a diverse set of 28 task-dataset pairs, employing instruction tuning in a multi-task fashion. Unlike MolT5 and Text+Chem T5, Nach0 employs separate tokenization for chemical atoms and natural language tokens. 

BioT5 \cite{pei2023biot5} utilizes custom tokenization for SELFIES and natural language sequences and is fine-tuned on 15 tasks related to molecule and protein property prediction, drug-target interaction, and protein-protein interaction. 
DrugChat \cite{liang2023drugchat} employs a GNN for encoding molecule graphs, a large LM (LLM), and an adaptor to convert graph representations for LLM compatibility. 
Following Transformer-M \cite{luo2022one}, MolLM \cite{tang2023mollm} offers a unified pre-training framework with a text Transformer encoder and a molecular Transformer encoder, pre-trained on molecular graphs, handling both 2D and 3D structures with attention mechanisms incorporating edge features and 3D spatial relationships.

\subsection{Linker generation}

Three models have emerged as significant contributors to linker generation: DeLinker\cite{imrie2020delinker}, 3DLinker\cite{huang223dlinker}, and DiffLinker\cite{igashov2022difflinker}.

DeLinker is a model adapted for linker design. It particularly retains the 3D structural information and generates linkers by providing two input fragments. It’s one of the first attempts to apply Graph Neural Networks (GNN) in linker design.

3DLinker is an E(3) Equivariant Variational Autoencoder for Molecular Linker Design. It generates a small “linker” to physically attach two independent molecules with their distinct functions. The generation of linkers is conditional on the two given molecules, and the linkers heavily depend on the anchor atoms of the two molecules to be connected. It predicts anchor atoms and jointly generates linker graphs and their 3D structures3.

DiffLinker is an Equivariant 3D-conditional Diffusion Model for Molecular Linker Design. Given a set of disconnected fragments in 3D, DiffLinker places missing atoms in between and designs a molecule incorporating all the initial fragments. Unlike previous approaches, which can only connect pairs of molecular fragments, DiffLinker can link an arbitrary number of fragments.

\subsection{LLaMa baseline}\label{app:llama}
For establishing a comparative single-domain baseline we trained the model and tokenizer from scratch with LLaMa-like architecture. For both cases, we adopted the implementations of Open LLaMa \cite{openlm2023openllama} from the HuggingFace framework and performed training on the GEOM dataset.

In this study, the Tokenizer remains consistent with the Open LLaMa version, though it was retrained solely on recent datasets. We made no improvements to the tokenization process for digits or molecules and restricted the vocabulary size to 512 tokens.

Concerning the model, modifications had been applied to its configuration to align with the dimensions of \ourmodel{}. Namely, we set the number of hidden layers and attention heads to 12, hidden size to 768, and intermediate size to 2048. The overall parameter count of the model reached nearly 350M. During the training procedure, we set the global batch size equal to 32, the learning rate to 1e-4 with a linear scheduler with warm-up steps, and weight decay to 1e-2. We trained our model using the causal language modeling objective for 10 epochs.

\subsection{Attention-based models for point clouds}

Recently, there have been several investigations into applying the Transformer architecture to point cloud analysis. The Point Cloud Transformer (PCT) \cite{guo2021pct} utilizes a permutation-invariant transformer instead of a self-attention mechanism to manage unstructured and disordered point data in irregular domains. Similarly, a Transformer-based network (TR-Net) 
PointConT \cite{liu2023point} leverages the locality of points in the feature space by clustering sampled points with similar features into the same class and computing self-attention within each class. The design is intended to capture long-range dependencies within the point cloud while maintaining computational efficiency. 
\cite{liu2022tr} employs a neighborhood embedding strategy along with a residual backbone featuring skip connections to enhance context-aware and spatial-aware features. The network utilizes an offset attention operator on point cloud spatial information to refine attention weights, thereby improving the extraction of global features.

\subsection{Non-diffusion approaches.} 

Several neural generative models were proposed to generate spatial molecular structures, including generators working directly with atomic density grid and voxels. A conditional variational autoencoder \cite{ragoza2022ligan} was trained on atomic density grid representations of cross-docked protein-ligand structures. To construct valid molecular conformations from the generated atomic densities, atom fitting and bond inference procedures were utilized. Pocket2Mol \cite{peng2022pockettomol}, an E(3)-equivariant generative network, consists of two modules: 1) a graph neural network (GNN) that captures both spatial and bonding relationships between atoms in the binding pockets, and 2) an algorithm that samples drug candidates based on pocket representations from a tractable distribution. VoxMol \cite{pinheiro2023voxmol} samples noisy density grids from a smooth distribution using underdamped Langevin Markov chain Monte Carlo and denoises the noisy grid in a single step to refine the exact atom positions. Unlike point-cloud diffusion models, VoxMol is simpler to train, does not require prior knowledge of the number of atoms, and does not treat features as different distributions.
\section{Point Cloud Encoder Details and Visualizations} \label{app:pc_details}

\begin{figure*}
\begin{center}
\centerline{\includegraphics[width=1.45\columnwidth]{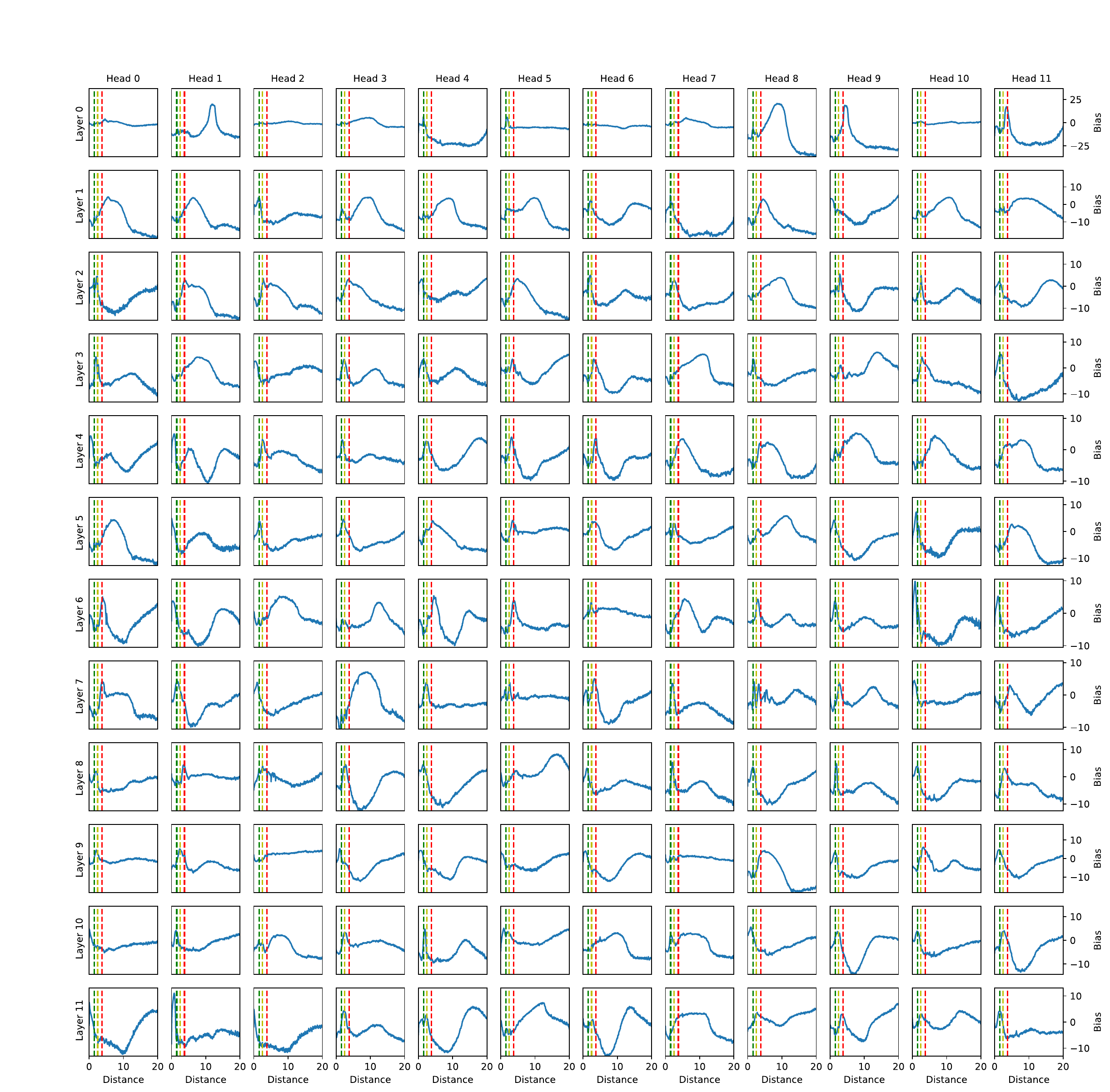}}
\caption{Visualization of learned relative biases. \textcolor{green}{Green} line corresponds to a typical C-C single covalent bond. \textcolor{yellow}{Yellow} line corresponds to typical hydrogen bond length. \textcolor{red}{Red} line corresponds to the typical distance between consequent CA-CA atoms in the protein.}
\label{fig:relative_bias}
\end{center}
\end{figure*}

\begin{figure*}
\begin{center}
\centerline{\includegraphics[width=0.84\columnwidth]{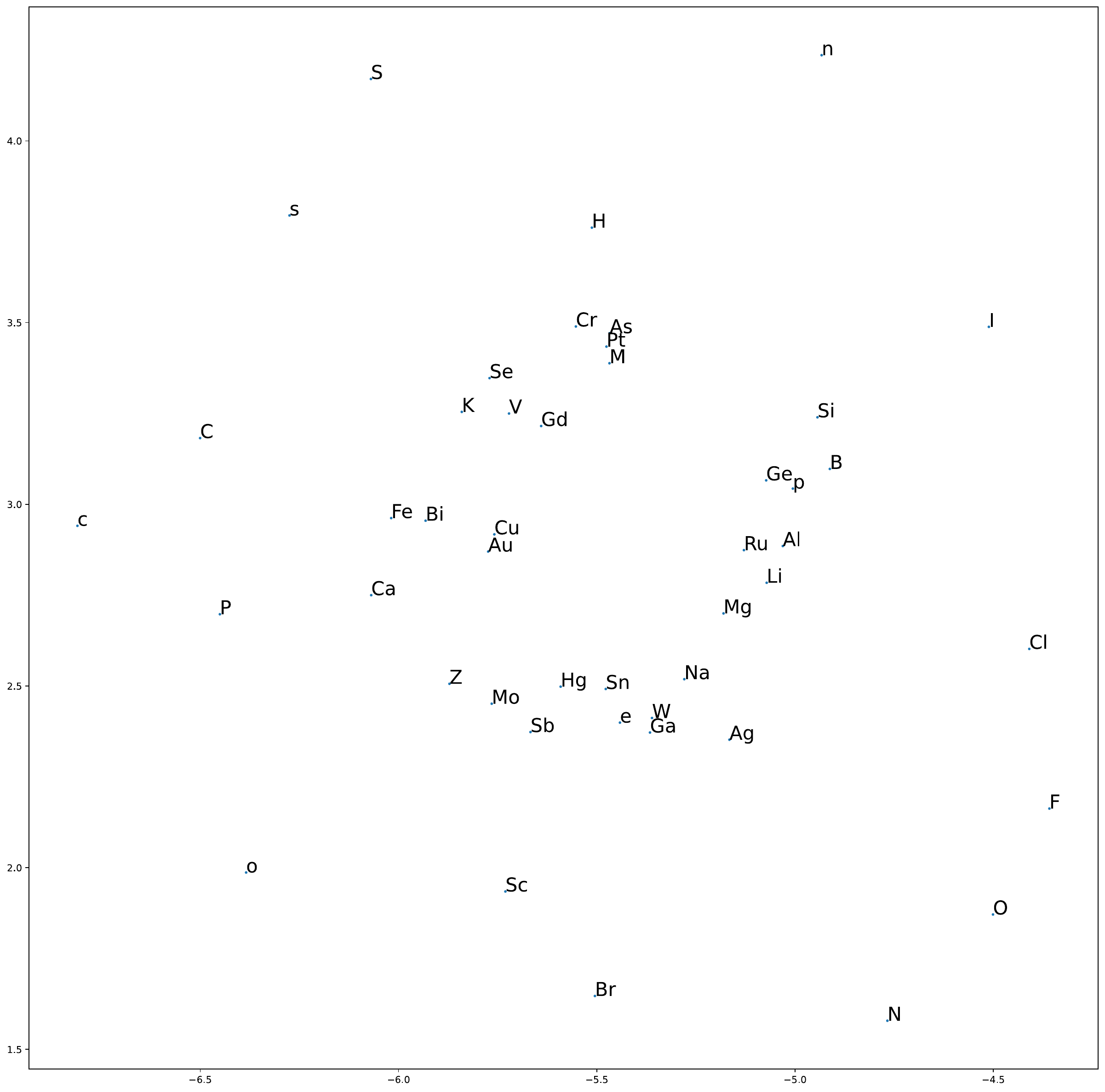}
\includegraphics[width=0.84\columnwidth]{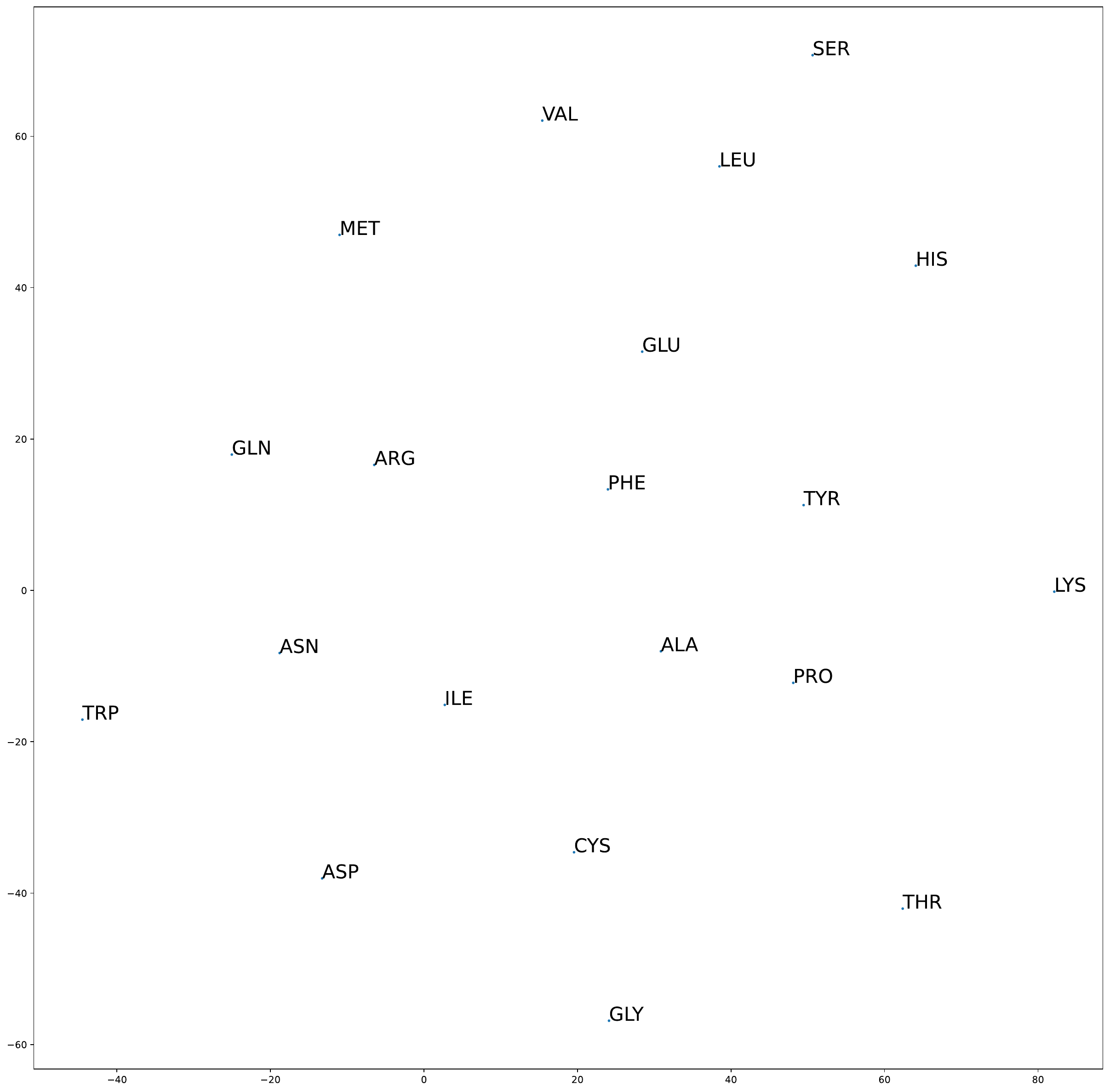}}
\caption{t-SNE visualization of learned atom (left) and amino acid (right) tokens embeddings.}
\label{fig:atom_embeddings}
\end{center}
\end{figure*}

Figure \ref{fig:relative_bias} illustrates the activation values of relative biases as a function of the distances between atoms. On the X-axis, the distances are represented, ranging from 0 to 20 \AA, while the Y-axis depicts the corresponding attention bias values. This visualization was conducted using synthetic data, where we systematically varied the distances between atom coordinates. The figure includes dashed lines that indicate typical atomic distances: the green dashed line represents the typical length of a C-C single covalent bond, the yellow dashed line corresponds to the typical hydrogen bond length, and the red dashed line signifies the typical distance between consecutive CA-CA atoms in a protein. Notably, the activations in some attention heads within particular layers reach their peak values precisely in the regions corresponding to these typical bond lengths, highlighting the model's sensitivity to biologically relevant atomic distances.

Figure \ref{fig:atom_embeddings} presents t-SNE visualizations of textual atom token embeddings and textual amino acid token embeddings. These embeddings are projected into a 2D space where similar tokens are positioned based on their common chemical and structural characteristics. For atom tokens, the visualization highlights how atoms with similar properties or roles within molecules are represented. Similarly, the amino acid token embeddings reflect their structural similarities, demonstrating the model's ability to capture and encode essential chemical and structural information within the textual token embeddings.

\section{Model Parameters and Training Details} \label{app:hyper}

In our research, we primarily utilize a model based on the nach0 \cite{nach0} architecture. Our experiments involve a \texttt{base} model variant, characterized by 12 layers, a hidden state of 768 dimensions, a feed-forward hidden state of 2048 dimensions, and 12 attention heads. We utilize the same parameters to build our point cloud encoder. The total model size has 370M parameters.

For the selected model, we conduct proposed pre-training and subsequent fine-tuning. The model was trained using two NVIDIA A6000 GPUs.

The pre-training and fine-tuning stages were executed using the following hyperparameters: a batch size of 64 for both pre-training and fine-tuning, a learning rate set to 1e-4, a weight decay of 0.01, and a cosine schedule. Both the pre-training and fine-tuning stages lasted for 100000 steps.
\section{Ablation study on design choices of \ourmodel{}} \label{app:analysis}

\begin{figure}
\begin{center}
\centerline{
\includegraphics[width=0.95\columnwidth]{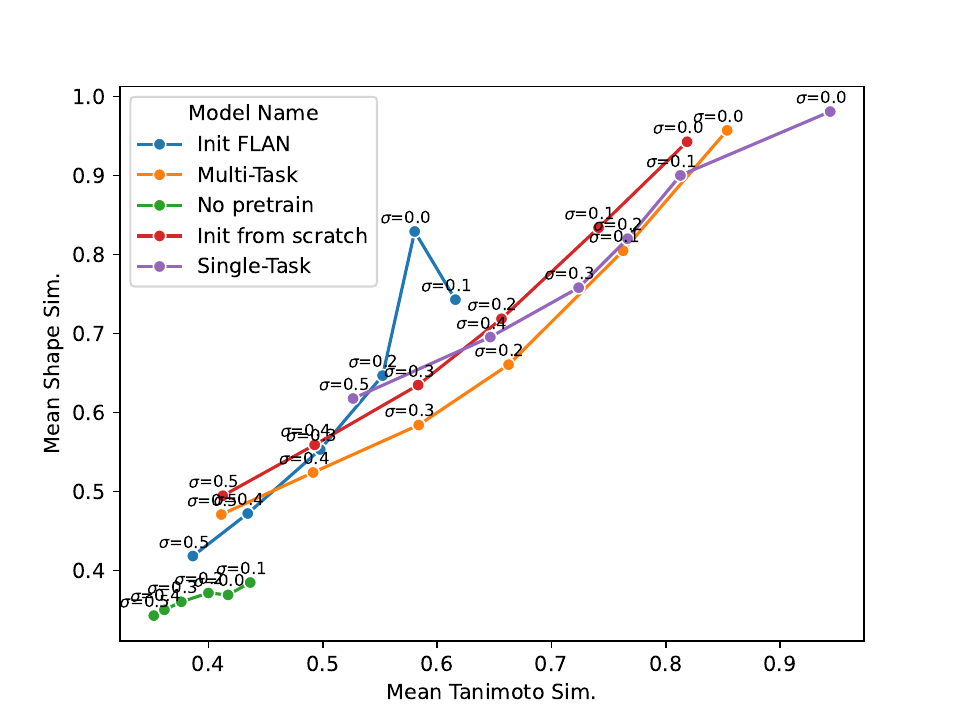}
}
\caption{Comparison of model performance on shape-conditioned generation. The figure evaluates \ourmodel{} in multi-task and single-task settings with nach0 backbone, as well as the performance with the FLAN and random LM component and no pre-train models. }
\label{fig:shape_conditioned_ablation}
\end{center}
\end{figure}

\begin{table}
\begin{center}
\begin{tabular}{| c | c | c | c | c | c | c | c | c | } 
 \hline
\multirow{3}{*}{Method} & \multicolumn{4}{c|}{Recall} & \multicolumn{4}{c|}{Precision} \\
\cline{2-9} 
  & \multicolumn{2}{c|}{COV ($\uparrow$)} & \multicolumn{2}{c|}{AMR ($\downarrow$)} & \multicolumn{2}{c|}{COV ($\uparrow$)} & \multicolumn{2}{c|}{AMR ($\downarrow$)} \\ 
 \cline{2-9} 
 & Avg & Med & Avg & Med & Avg & Med & Avg & Med  \\
 \hline
 Multi-Task & 50.1 & 50.0 & 0.76 & 0.75 & 27.7 & 16.7 & 1.11 & 1.11 \\
 \hline
 Single-Task & 57.7 & 59.5 & \textbf{0.70} & \textbf{0.69} & 32.4 & \textbf{23.1} & 1.03 & 1.03 \\
 \hline
 \multicolumn{9}{|c|}{Another init checkpoints} \\
 \hline
 FLAN & 38.7 & 26.7 & 0.85 & 0.85 & 20.0 & 8.3 & 1.21 & 1.19 \\
 \hline
 From scratch & 34.4 & 22.3 & 0.92 & 0.90 & 16.5 & 6.7 & 1.39 & 1.37 \\
 \hline
 \multicolumn{9}{|c|}{Input SMILES format} \\
 \hline
 Canonical & 44.2 & 39.5 & 0.82 & 0.81 & 25.9 & 14.3 & 1.15 & 1.15 \\
 \hline
 Non-isomeric & 50.4 & 47.3 & 0.76 & 0.76 & 27.3 & 16.7 & 1.12 & 1.12 \\
 \hline
 \multicolumn{9}{|c|}{Pre-training} \\
 \hline
 No pre-training & \textbf{58.0} & \textbf{60.6} & \textbf{0.70} & \textbf{0.69} & \textbf{34.2} & \textbf{25.0} & \textbf{1.02} & \textbf{1.02} \\ 
 \hline
\end{tabular}
\end{center}
\caption{Comparison of model performance on conformation generation. The table evaluates \ourmodel{} in multi-task and single-task settings with nach0 backbone, as well as the performance with the FLAN and random LM component and no pre-train models.  The better scores across all models are highlighted in bold.}
\label{tab:conf_gen_ablation}
\end{table}

\begin{table}
\begin{center}
\begin{tabular}{| c | c | c | c | c | c |} 
 \hline
 Method & Val. ($\uparrow$) & Uniq. ($\uparrow$) & Filt. ($\uparrow$)  & RMSD ($\downarrow$) & $SC$ ($\uparrow$) \\
 \hline
 Multi-Task & 81.6\% & 27.6\% & 99.00\% & 1.28 & 0.86 \\
 Single-Task & \textbf{89.7\%} & 12.3\% & \textbf{99.55}\% &  \textbf{1.04} & \textbf{0.88} \\
 \hline
 \multicolumn{6}{|c|}{Another init checkpoints} \\
 \hline
 FLAN & 81.1\% & 30.9\% & 98.97\% &  1.25 & 0.86 \\
 From scratch & 83.9\% & 24.1\% & 99.20\% & 1.17 & 0.87 \\
 \hline
 \multicolumn{6}{|c|}{Pre-training} \\
 \hline
 No pre-training & 34.8\% & \textbf{61.8\%} & 99.03\% & 1.59 & 0.83 \\
 \hline
\end{tabular}

\end{center}
\caption{Comparison of model performance on linker design. The table evaluates \ourmodel{} in multi-task and single-task settings with nach0 backbone, as well as the performance with the FLAN and random LM component and no pre-train models. The better scores across all models are highlighted in bold.}
\label{tab:linker_design_ablation}
\end{table}

\begin{table}
\begin{center}
\begin{tabular}{| c | c | c | c | c | c |} 
 \hline
 \multirow{2}{*}{Model} & \multirow{2}{*}{Valid. ($\uparrow$)} & \multirow{2}{*}{Div. ($\uparrow$)} & \multicolumn{2}{c|}{Vina Dock ($\downarrow$)} & \multirow{2}{*}{\makecell{High \\ Affinity ($\uparrow$)}}\\
 \cline{4-5}
& & & Avg. & Med &  \\
 \hline 
 Multi-Task & 91.78\% & 0.32 & -6.52 & \textbf{-6.86} & 38.2\% \\
 Single-Task & 89.82\% & \textbf{0.40} & -6.50 & -6.62 & \textbf{41.1}\% \\
 \hline
 \multicolumn{6}{|c|}{Another init checkpoints} \\
 \hline
 FLAN & 86.03\% & 0.34 & -6.26 & -6.67 & 36.1\% \\
 From scratch & \textbf{93.02}\% & 0.34 & \textbf{-6.77} & -6.82 & 38.4\% \\
 \hline
 \multicolumn{6}{|c|}{Pre-training} \\
 \hline
 No pre-training & \textbf{93.02}\% & 0.26 & -5.70 & -6.43 & 33.1\% \\
 \hline
\end{tabular}

\end{center}
\caption{Comparison of model performance on pocket-conditioned generation task. The table evaluates \ourmodel{} in multi-task and single-task settings with nach0 backbone, as well as the performance with the FLAN and random LM component and no pre-train models. The better scores across all models are highlighted in bold.}
\label{tab:pocket_conditioned_ablation}
\end{table}

\begin{table*}[t]
\begin{center}
\begin{tabular}{ | c | c | c | c || c | c || c |} 
 \hline
 \multirow{2}{*}{Group} & \multirow{2}{*}{Metrics}  &  \multicolumn{5}{c|}{\ourmodel{}} \\ 
 \cline{3-7} 
  & & Multi-Task & Single-Task & FLAN & From scratch & No pre-train \\
 \hline
 \multirow{3}{*}{Druglikeness} & QED ($\uparrow$) & 0.770 & 0.664 & 0.634 & 0.712 & \textbf{0.828} \\ 
  & SA ($\uparrow$) & \textbf{0.872} & 0.848 & 0.825 & 0.858 & \textbf{0.871} \\
  & Lipinski ($\uparrow$) & 4.993 & 4.938 & 4.923 & 4.973 & \textbf{5.0} \\  
 \hline
\multirow{3}{*}{3D structures} & JS. bond lengths ($\downarrow$) & 0.205 & \textbf{0.146} & 0.397 & 0.190 & 0.326 \\
 & JS. bond angles ($\downarrow$) & 0.107 & \textbf{0.100} & 0.247 & 0.137 & 0.152 \\ 
 & JS. dihedral angles ($\downarrow$) & 0.133 & \textbf{0.113} & 0.229 & 0.129 & 0.188 \\
 \hline
\multirow{4}{*}{Bonds} & JS. num. bonds per atoms ($\downarrow$) & 0.230 & \textbf{0.094} & 0.181 & 0.159 & 0.537 \\
& JS. freq. bond types  ($\downarrow$) & 0.050 & \textbf{0.033} & 0.067 & 0.061 &  0.095 \\
& JS. freq. bond pairs  ($\downarrow$) & 0.038 & \textbf{0.033} & 0.101 & 0.040 & 0.056 \\
& JS. freq. bond triplets  ($\downarrow$) &  0.054 & \textbf{0.041} & 0.123 & 0.043 & 0.085 \\
\hline
\multirow{3}{*}{Rings} & JS. num. rings ($\downarrow$) & 0.267 & \textbf{0.036} & 0.079 & 0.154 & 0.492 \\
 & JS. num. n-sized rings ($\downarrow$) & 0.059 & \textbf{0.024} & 0.045 & 0.038 & 0.102 \\
 & Num. Intersecting rings ($\uparrow$) & \textbf{9} & 8 & 6 & \textbf{9} & 6 \\ 
\hline
Mean RMSD & Mean RMSD ($\downarrow$) & 1.087 &	1.347 &	1.492 &	1.259 &	\textbf{0.832} \\
\hline

\end{tabular}
\end{center}
\caption{Comparison of model performance on distribution learning. The table evaluates \ourmodel{} in multi-task and single-task settings with nach0 backbone, as well as the performance with the FLAN and random LM component and no pre-train models. The better scores across all models are highlighted in bold.}
\label{tab:distribution_learning_3d_ablation}
\end{table*}

This section explores the critical role of the language model (LM) within \ourmodel{}. It begins with selecting the LM component, where \ourmodel{} is initialized with nach0, a top-tier chemical LM. Comparative analyses reveal that models leveraging nach0 outperform those using other LMs across various tasks, including distribution learning, conformation generation, linker design, and pocket-conditioned generation. Notably, nach0-based models consistently exhibit superior performance, underscoring the efficacy of domain-specific pre-training. This section provides more details of design choices of \ourmodel{} on spatial molecular distribution learning (Tab. \ref{tab:distribution_learning_3d_ablation}) , conformation generation (Tab. \ref{tab:conf_gen_ablation}), linker design (Tab. \ref{tab:linker_design_ablation}), shape-conditioned generation (Fig. \ref{fig:shape_conditioned_ablation}) and pocket-conditioned generation (Tab. \ref{tab:pocket_conditioned_ablation}) tasks.

\subsection{LM component}

The first important design choice of \ourmodel{} is the LM component. For all experiments presented in the main part of our paper, we initialize \ourmodel’s LM component with nach0 \cite{nach0}, the state-of-the-art chemical LM for multi-domain tasks. In this section, we compare several LMs: (i) the state-of-the-art cross-domain nach0 \cite{nach0}, (ii) the general-domain FLAN \cite{flant5}, and (iii) random initialization.

As for the distribution learning task, several observations can be made from Table \ref{tab:distribution_learning_3d_ablation}. First, the single-task and multi-task \ourmodel{} with nach0 outperformed other models on 3D Structures, Bond, and Ring Metrics. Second, the single-task model has the lowest JS divergence (0.146), followed by the multi-task model (0.205). The single-task model performs best (0.100 in the JS bond angles metric), while the multi-task model is close behind (0.107). The FLAN-based model shows the highest divergence (0.247). Overall, the single-task model consistently achieves the lowest JS divergences across various metrics, indicating high accuracy in bond and ring structures. Both \ourmodel{} initialized with nach0 and the random-based LM components outperform the general-domain LM component FLAN. This indicates that \ourmodel{} with domain-specific nach0 provides significant improvements over the model with general-domain FLAN in downstream tasks in both single-task and multi-task settings. 

Similar observations can be made for the conformation generation task. Table \ref{tab:conf_gen_ablation} indicates that incorporating multi-tasking and domain-specific pre-training significantly enhances the quality metrics for the task of generating conformations.

In addition to the improvements observed in the task of generating conformations, the table \ref{tab:linker_design_ablation} highlights the substantial benefits of incorporating multi-tasking and domain pre-training for the linker design task. Similar to the enhancements seen in conformation generation, the inclusion of multi-task learning and pre-training techniques significantly boosts the quality metrics for linker design.

Corresponding outcomes emerge when examining the results of the pocket-conditioned generation task. The integration of multi-tasking and pre-training methods leads to noticeable improvements in quality metrics \ref{tab:pocket_conditioned_ablation}.

Another way to compare the impact of implemented enhancements is to analyze the dependence curve, which delineates the relationship between Mean Tanimoto similarity and Mean shape similarity across various alpha values. A robust model is expected to yield lower Tanimoto similarities for the same levels of shape similarity. Figure \ref{fig:shape_conditioned_ablation} illustrates these curves for models with different levels of ablations.  The figure demonstrates that the multi-task model performs worse than the single-task model, as evidenced by its higher Tanimoto similarities for comparable shape similarities. 

\begin{table*}[t]
\begin{center}
\begin{tabular}{|p{2.5cm}|p{3cm}|p{2cm}|p{1.5cm}|p{1cm}|p{1.5cm}|}\hline
Model & Task & Time, hours ($\downarrow$) & CO2 Emissions, kg. CO2 eq. ($\downarrow$) & Total time, hours ($\downarrow$) & Total CO2 Emissions, kg. CO2 eq. ($\downarrow$) \\\hline
\multirow{2}{*}{MolDiff} & Training & 60 & 5.55 & \multirow{2}{*}{180} & \multirow{2}{*}{16.65} \\
 & 10k Sampling & 120 & 11.1 &  &  \\\hline
\multirow{2}{*}{EDM} & Training & 730 & 67.53 & \multirow{2}{*}{840} & \multirow{2}{*}{77.71} \\
 & 10k Sampling & 110 & 10.18 &  &  \\\hline
\multirow{5}{*}{\textbf{\ourmodel{}}} & Pre-train & 39.5 & 4.98 & \multirow{4}{*}{164.5}  
& \multirow{4}{*}{20.73}
\\
 & Finetune (multi-task, 2 GPU) & 119 & 14.99 &  &  \\
 & Finetune (single task, 1 GPU)
 & 59.5 & 7.5 &  &  \\
 & 10k Sampling & 6 & 0.76 &  &  \\
& Resources per 1 task & - & - &  \textbf{27.4} &   \textbf{3.5}\\\hline
\multirow{2}{*}{Tor. Diff. (*)} & Training & 360 & 33.72 & \multirow{2}{*}{393} & \multirow{2}{*}{38.33} \\
 & 10k Sampling & 33 & 4.61 &  &  \\\hline
\multirow{2}{*}{TargetDiff (*)}
& Training & 24 & 2.9 & \multirow{2}{*}{95} & \multirow{2}{*}{12.23} \\
 & 10k Sampling & 71 & 9.33 &  &  \\\hline
\multirow{2}{*}{Pocket2Mol (*)} & Training & 132 & 11.8 & \multirow{2}{*}{227} & \multirow{2}{*}{26} \\
 & 10k Sampling & 95 & 14.2 &  & \\\hline
\end{tabular}
\end{center}
\caption{GPU computation time and CO2 emissions for \ourmodel{} and state-of-the-art diffusion models; timings, marked with *, are extracted from author's papers. Single-task fine-tuning was excluded from the total time and total CO2 emission for \ourmodel{} and written here only for the direct comparison purpose with single-task models}
\label{tab:co2_emission}
\end{table*}

\subsection{Impact of 3D pre-training}
Another significant contribution of \ourmodel{} is its 3D pre-training approach. Taking into account the three fundamental aspects of machine learning -- data and task complexity -- pre-training is advantageous when the amount of data available for the downstream task is relatively small compared to the complexity of the task. 

As shown in Table \ref{tab:distribution_learning_3d_ablation}, 
This indicates that pre-trained and fine-tuned \ourmodel{} provides significant improvements over fine-tuned \ourmodel{} without pre-training 
on 3D Structures, Bond, and Ring Metrics. Surprisingly, model pre-training achieved better results than single-task \ourmodel{} on the conformation generation task. As for the linked design task, we can make similar observations as on the distribution learning task: 3D pre-training helps the model perform the downstream task.

\subsection{Canonical and non-isometric SMILES} 
\ourmodel{} augments isomeric SMILES that include stereolabels. This allows the model to utilize some 3D information present in stereolabels of isomeric SMILES and to generalize better with augmentation. We conduct ablation studies to compare this choice against a non-augmented version of isomeric SMILES, which is called canonical SMILES in Table \ref{tab:conf_gen_ablation}, as well as non-isomeric SMILES without augmentation. Table \ref{tab:conf_gen_ablation} shows that the main multi-task \ourmodel{} model outperforms the canonical isomeric SMILES approach without augmentation on conformation generation but is on par with the non-augmented non-isomeric SMILES approach. The results suggest that augmentation boosts performance. However, a non-isomeric SMILES choice might also be an option for unconditional conformation generation with relatively large and diverse datasets.

\section{Computational Resources and Training Time}\label{app:compres}
\subsection{Hardware Computational Resources}

We utilized two NVIDIA RTX A6000 GPUs with 48 GB of memory, CPU with 60 computational cores and 128 GB of RAM for \ourmodel{} training.

\paragraph{Pre-training Time:} The initial pre-training phase took 40 hours. 

\paragraph{Finetuning time:} The Finetuning phase took around 60 hours.

\paragraph{Evaluation Time:} The evaluation phase, which involved running inference, calculating performance metrics, and validating results, took an additional 6 hours (excluding the ablation study sampling).
\section{Model Training Time and CO2 Impact}\label{app:co2}

In this section, we have analyzed the computational requirements and associated carbon dioxide (CO2) emissions for training our proposed model and state-of-the-art diffusion models discussed in this article. The detailed results are presented in Table \ref{tab:co2_emission}.

All experiments were conducted utilizing the CoreWeave infrastructure. For our model, denoted as \ourmodel{}, the training and evaluation were performed on an Nvidia RTX A6000 GPU with 48GB of memory and a thermal design power (TDP) of 300W. The total training and evaluation time for our model was 164.5 hours, resulting in an estimated CO2 emission of 20.73 kgCO2eq. For the training and evaluation of MolDiff and EDM models, we utilized an Nvidia RTX A4000 GPU with 16GB of memory and a TDP of 140W. The total GPU time required for these models was 1020 hours, leading to an estimated CO2 emission of 94.36 kgCO2eq. These estimations were conducted using the Machine Learning Impact calculator, as presented in \cite{lacoste2019quantifying}. CO2 values and GPU hours for TorDiff, TargetDiff and Pocket2Mol were extracted from their respective original works.

From Table \ref{tab:co2_emission}, it can be observed that the full training cycle for our \ourmodel{} model requires 164.5 hours, which is the second-best result after TargetDiff, which requires substantially less training time of 95 GPU hours. In terms of CO2 emissions, our model ranks third after TargetDiff and MolDiff. However, a key advantage of our model over other state-of-the-art models is its multitasking capability. When considering the GPU time spent per task, our model requires only 27.4 hours, which is significantly better than TargetDiff. Furthermore, considering the CO2 emissions per single task, our model requires only 3.5 kg CO2eq of emissions. Thus, our model is much more effective and efficient compared to other state-of-the-art models in terms of computational requirements and environmental impact. 
\section{Licenses for Existing Assets}\label{app:license}
We used the following datasets: 1) the GEOM dataset released under the CC0 1.0 license, 2) the MOSES dataset released under MIT License, 3) The ZINC dataset released under MIT License, and the CrossDocked dataset released under CC0 1.0 Universal (CC0 1.0) Public Domain Dedication.
We also used the following models: 1) Nach0 was released under Creative Commons Attribution Non-Commercial 4.0, and 2) The FLAN model was released under the Apache 2.0 License.
To perform ablation studies, we have used these model architectures: 1)OpenLLaMa source code is released under the Apache License, and 2) MolDiff model source code is released under the MIT License.

\end{document}